\documentclass[acmsmall]{acmart}
\usepackage{array}
\usepackage{longtable}
\usepackage{graphicx}
\usepackage{makecell}
\usepackage{booktabs}
\usepackage{float}
\AtBeginDocument{%
  }

\setcopyright{acmlicensed}
\copyrightyear{2024}
\acmYear{2024}
\acmDOI{XXXXXXX.XXXXXXX}

\begin{document}

\title{A Survey on Deep Tabular Learning}


\author{Shriyank Somvanshi}
\affiliation{%
  \institution{Texas State University}
  \city{San Marcos}
  \country{TX}}
\email{jum6@txstate.edu}

\author{Subasish Das, Ph.D.}
\affiliation{%
  \institution{Texas State University}
  \city{San Marcos}
  \country{TX}}
\email{subasish@txstate.edu}

\author{Syed Aaqib Javed}
\affiliation{%
  \institution{Texas State University}
  \city{San Marcos}
  \country{TX}}
\email{aaqib.ce@txstate.edu}

\author{Gian Antariksa, Ph.D.}
\affiliation{%
  \institution{Texas State University}
  \city{San Marcos}
  \country{TX}}
\email{gian.antariksa@gmail.com}

\author{Ahmed Hossain, Ph.D.}
\affiliation{%
  \institution{Texas State University}
  \city{San Marcos}
  \country{TX}}
\email{ahmed.hossain@txstate.edu}

\renewcommand{\shortauthors}{Somvanshi et al.}

\begin{abstract}
  Tabular data, widely used in industries like healthcare, finance, and transportation, presents unique challenges for deep learning due to its heterogeneous nature and lack of spatial structure. This survey reviews the evolution of deep learning models for tabular data, from early fully connected networks (FCNs) to advanced architectures like TabNet, SAINT, TabTranSELU, and MambaNet. These models incorporate attention mechanisms, feature embeddings, and hybrid architectures to address tabular data complexities. TabNet uses sequential attention for instance-wise feature selection, improving interpretability, while SAINT combines self-attention and intersample attention to capture complex interactions across features and data points, both advancing scalability and reducing computational overhead. Hybrid architectures such as TabTransformer and FT-Transformer integrate attention mechanisms with multi-layer perceptrons (MLPs) to handle categorical and numerical data, with FT-Transformer adapting transformers for tabular datasets. Research continues to balance performance and efficiency for large datasets. Graph-based models like GNN4TDL and GANDALF combine neural networks with decision trees or graph structures, enhancing feature representation and mitigating overfitting in small datasets through advanced regularization techniques. Diffusion-based models like the Tabular Denoising Diffusion Probabilistic Model (TabDDPM) generate synthetic data to address data scarcity, improving model robustness. Similarly, models like TabPFN and Ptab leverage pre-trained language models, incorporating transfer learning and self-supervised techniques into tabular tasks. This survey highlights key advancements and outlines future research directions on scalability, generalization, and interpretability in diverse tabular data applications.
\end{abstract}




\received{October 10, 2024}

\maketitle

\section{Introduction}
Tabular data, which consists of rows and columns representing structured information \cite{ucar2021subtab, zhu2021converting}, is the most commonly used data format in many industries, including healthcare, finance, and transportation. Unlike unstructured data such as images and text, tabular data directly represents real-world phenomena in a structured form, making it crucial for decision-making processes in areas like risk assessment, predictive analytics, and safety monitoring. For example, in the field of transportation engineering, tabular data plays a key role in recording crash incidents, vehicle attributes, environmental factors, and human behavior, enabling researchers to predict crash severity and improve safety measures using data-driven insights.

Despite the success of deep learning in domains like computer vision and natural language processing (NLP), its application to tabular data has been less straightforward. Deep learning models often struggle with tabular data due to several challenges:

\begin{enumerate}
  \item \textbf{Small sample sizes:} Many tabular datasets are relatively small, especially when compared to large image or text datasets, leading to overfitting in complex deep learning models.
  \item \textbf{High dimensionality:} Tabular data often involves many features, which can be sparse or irrelevant, making it difficult for models to identify meaningful patterns.
  \item \textbf{Complex feature interactions:} Unlike images or text, where local structures are prominent, the interactions between features in tabular data are non-local and complex, requiring more specialized architectures to capture these relationships effectively.
\end{enumerate}

These factors make tree-based models like \texttt{XGBoost} and \texttt{Random Forests} more effective for many tabular data tasks, as they are better suited to handle sparse features and complex interactions. Over recent years, significant strides have been made in the development of deep learning models specifically for tabular data, addressing the unique challenges posed by this data type. While early models like fully connected networks (FCNs) showed promise, new architectures have emerged that have significantly advanced the field \cite{ahamed2024mambatab, borisov2022language, chen2023recontab, wu2024switchtab}. One of the leading models in this space is the FT-Transformer, which adapts transformer models, initially developed for sequential data, to effectively handle tabular data by encoding features through attention mechanisms \cite{borisov2022deep, shwartz2022tabular}. This model has shown impressive performance due to its ability to learn complex interactions between features, making it well-suited for high-dimensional data.

Another recent innovation is the Self-Attention and Intersample Transformer (SAINT), which improves upon the original transformer by introducing intersample attention mechanisms, allowing the model to better capture relationships between rows of tabular data \cite{grinsztajn2022tree}. SAINT has demonstrated superior performance in various benchmarks compared to traditional models like XGBoost and deep learning models such as Neural Oblivious Decision Ensembles (NODE). Additionally, models like TabTransformer leverage transformers specifically for categorical feature encoding, providing a more scalable solution for handling mixed data types in tabular datasets. This approach enables the model to capture meaningful representations from categorical variables, which are often challenging for traditional deep-learning architectures to handle effectively. These new models have introduced significant innovations in terms of feature encoding, complex interaction learning, and model interpretability, which are crucial for advancing the application of deep learning to tabular data in many research areas. The objective of this survey paper is to review these advancements in detail, exploring their historical evolution as seen in Figure 1, key techniques, datasets, and potential applications. 

\begin{figure}[htp]
    \centering
    \includegraphics[width=0.98\linewidth]{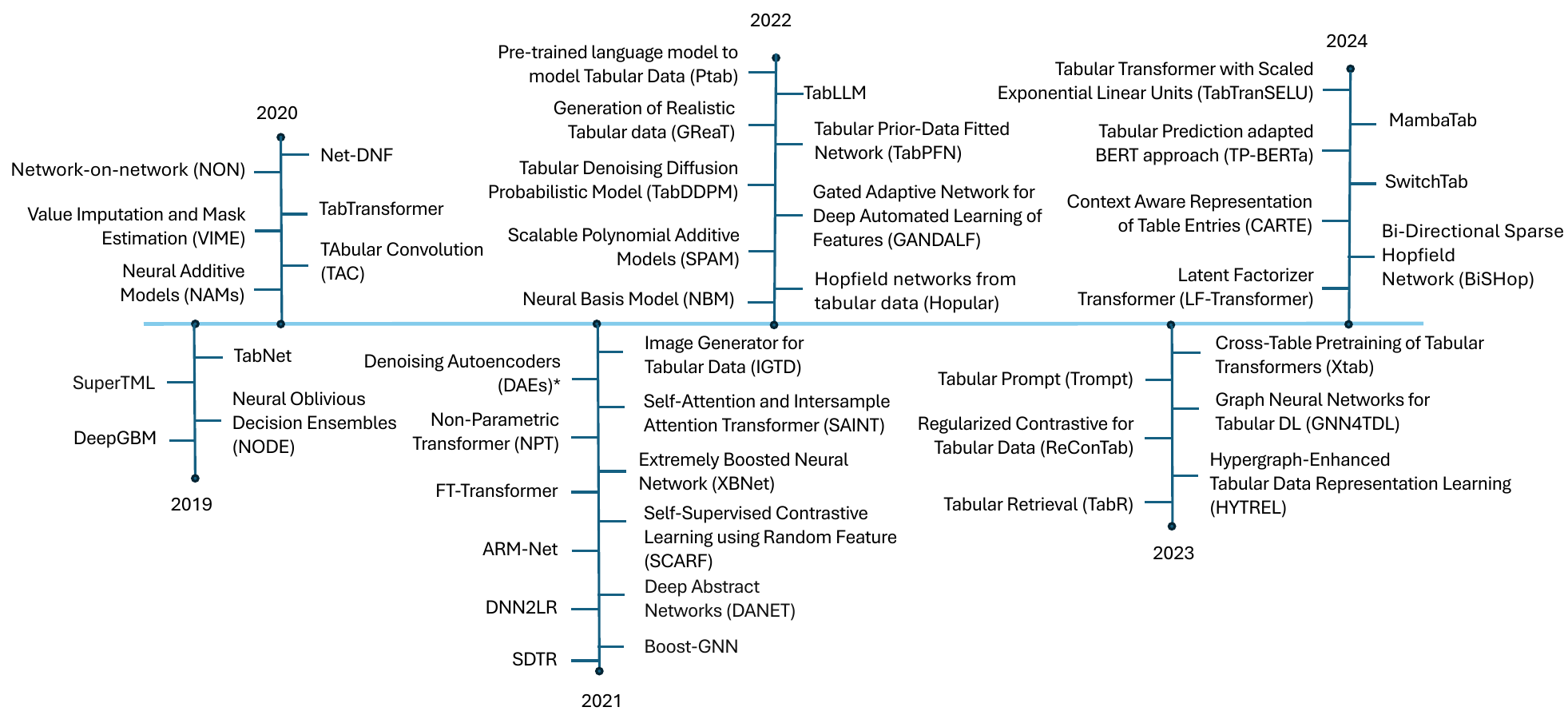}
    \caption{Progression of Tabular Deep Learning Models}
    \label{fig:Progression TDL}
\end{figure}

\section{Challenges in Modeling Tabular Data}
\subsection{Heterogeneous Feature Types}
Tabular data, a foundational structure in fields like healthcare, finance, and transportation, commonly contains heterogeneous data types such as numerical, categorical, ordinal, text, and even multimedia elements like images and emojis. Numerical features often represent continuous or discrete values (e.g., age, income), while categorical features classify entities into discrete groups (e.g., gender, city) \cite{wu2024deep, gorishniy2022embeddings}. In more complex cases, text data, images, or emojis may be embedded within tables, providing rich context but complicating feature representation and model training. Understanding how to handle these varied data types is critical to improving the performance of deep learning models on tabular data.

Tabular data can also be represented in two different formats: 1D tabular data, and 2D tabular data as shown in Figure 2 below. In 1D tabular data, each row represents a sample and columns represent specific features, making it easy to process and analyze. This format is ideal for traditional machine learning tasks, as each column follows a specific data type and the structure is fixed. For example, in transportation safety datasets, each row could represent an individual crash event, and the columns might include features like vehicle speed, crash time, or road conditions. The simplicity of this structure makes it highly useful in various fields.

\begin{figure}[h!]
\centering
\includegraphics[width=1\columnwidth]{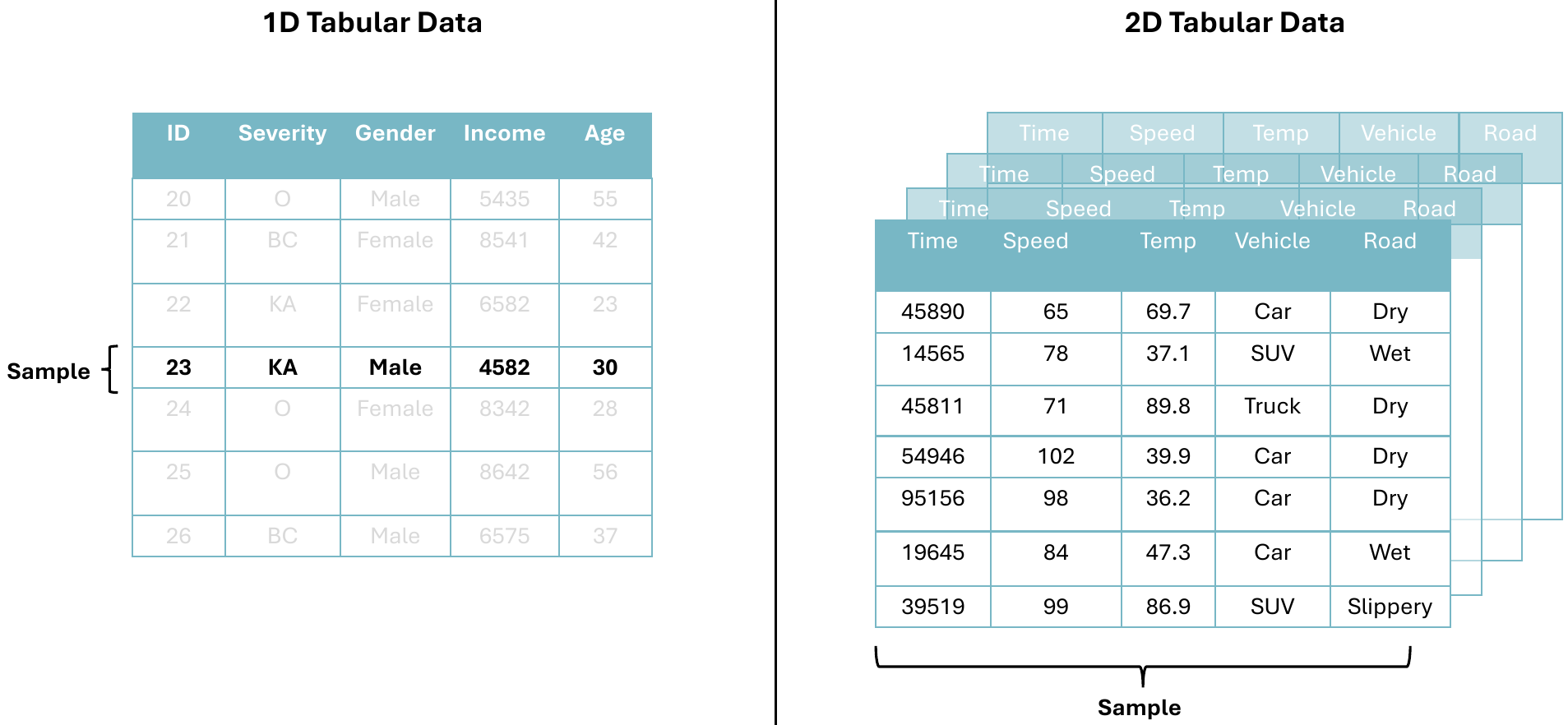}

\caption{An Illustration of 1D (left) and 2D (right) Tabular Dataset}

\label{fig:flow}
\end{figure}

In contrast, 2D tabular data provides a more complex format where each sample can be represented by a table, with multiple rows and columns within each table. This format is often used for tasks that require deeper relational analysis, such as tracking patient health over time or analyzing transportation data across different regions and times. 2D tabular data is also more flexible, incorporating diverse data types, including timestamps or unstructured data, like text or images, within each table. This additional complexity makes it suitable for applications in areas such as healthcare and transportation, where temporal and multi-dimensional data are critical. 

Understanding how to handle these varied data types is critical to improving the performance of deep learning models on tabular data. Some of these data types are explained below:

\begin{itemize}
\item {\texttt{Binary Data}}: Binary data, a type of categorical data with two possible values (such as, "Yes/No"), is often represented as 0 or 1 in deep learning models \cite{collett2002modelling}.
\item {\texttt{Numerical Data}}: Numerical data, representing continuous or discrete variables (e.g., age, vehicle speed), is common in predictive modeling, especially in transportation safety \cite{zeger1986longitudinal}. Deep learning models handle it directly, but preprocessing, like scaling or standardization, is critical for performance. Advanced techniques, such as numerical embeddings, help capture non-linear relationships and interactions in the data.
\item {\texttt{Timestamps}}: Timestamps provide essential temporal information in systems like traffic management. Preprocessing involves extracting features such as day, month, or hour to capture temporal patterns for deep learning models \cite{dyreson1993timestamp}.
\item {\texttt{Text Data}}: Text data in tabular formats, such as crash descriptions, presents challenges for deep learning models. Methods like TF-IDF \cite{hakim2014automated} and word embeddings (e.g., word to vector, global vectors for word representation) convert text into numerical vectors \cite{mikolov2013efficient, pennington2014glove}. Advanced models like transformers (e.g., BERT) capture context-aware embeddings \cite{li2020enhancing}.
\item {\texttt{Image Data}}: In multi-modal datasets, image data is sometimes embedded in tables, such as in autonomous driving, where road images are paired with tabular data. Convolutional Neural Networks (CNNs) process images, but integrating image features with tabular data requires feature fusion techniques. Hybrid models like TabTransformer use attention mechanisms to merge image and tabular data, enhancing predictive performance \cite{huang2020tabtransformertabulardatamodeling}.
\item {\texttt{Hyperlinks}}: Hyperlinks, though uncommon in traditional tabular datasets, are increasingly used in web data applications or web documents \cite{ma2021pre}. When tables include URLs, advanced preprocessing is required to extract metadata or context from the linked pages, often using NLP models to incorporate this information into the feature set.
\item {\texttt{Video Data}}: Video data in tabular formats provides valuable temporal information for domains like autonomous driving and traffic management. Keyframes from videos are processed using 3D-CNNs or Recurrent Neural Networks (RNNs) to capture spatial and temporal features, which are then integrated with tabular data to improve model predictions, such as in crash prediction models where video features enhance the understanding of road conditions and driver behavior \cite{lu2020new, maaloul2017adaptive}.
\item {\texttt{Emoji}}: Emojis common in social media and messaging platforms, enhance communication by visually conveying emotions or objects \cite{kollipara2021emoji} and pose challenges for encoding sentiment. Deep learning models use character-level or emoji embeddings to map them to sentiment vectors, enabling effective interpretation alongside other data types.

\end{itemize}

Tabular data, composed of rows and columns, lacks the spatial or sequential structure found in images and text, making it difficult to apply traditional deep learning models like CNNs, which rely on spatial coherence. Unlike structured data, reordering columns or rows in tabular data doesn't change feature relationships, and deep learning models struggle without the inductive biases that machine learning models like XGBoost and Random Forests possess. Machine learning models excel in handling heterogeneous feature types, non-local interactions, and small, high-dimensional datasets, where deep learning models often overfit and fail to generalize.

To address the limitations of traditional deep learning models when applied to tabular data, recent advancements have led to the development of specialized architectures such as TabNet, TabTransformer, and SAINT. These models introduce mechanisms like attention layers, feature embeddings, and hybrid architectures to dynamically focus on the most relevant features, improving their ability to handle the complexity of heterogeneous tabular data. For instance, TabNet \cite{arik2021tabnet} employs sequential attention mechanisms for instance-wise feature selection, while TabTransformer \cite{huang2020tabtransformertabulardatamodeling} uses self-attention layers to capture feature dependencies more effectively than CNNs. SAINT \cite{somepalli2021saint} enhances this approach by incorporating intersample attention, enabling the model to capture relationships between data rows. Moreover, models such as TabTranSELU \cite{mao2024tabtranselu} and GNN4TDL \cite{li2023graph} are designed to efficiently manage both categorical and numerical features by employing hybrid structures and regularization techniques, which help mitigate overfitting and improve generalization. These innovations have enabled deep learning models to rival or surpass traditional machine learning methods in tasks involving tabular data, including fraud detection, and predictive analytics. Additionally, novel techniques such as transforming tabular data into image-like structures \cite{zhu2021converting, somvanshi4847615enhanced}, employing multi-view representation learning, and extracting schemas from tabular data \cite{adelfio2013schema} further contribute to overcoming the challenges posed by the absence of inherent spatial relationships in tabular datasets.

By leveraging these advancements, recent tabular deep learning models not only address the unique challenges of tabular data but also offer significant improvements in performance, interpretability, and scalability compared to both traditional deep learning and machine learning approaches. These innovations demonstrate the growing potential for deep learning in handling complex, non-spatial data across a wide range of real-world applications.

\subsection{Non-spatial relationships}

Traditional deep learning models, such as CNNs and RNNs, excel in capturing spatial and sequential relationships in structured data types like images and text, where spatial coherence or temporal dependencies play a crucial role. CNN, for instance, detects local patterns by processing spatially adjacent pixels, allowing them to capture meaningful features through convolutions \cite{alzubaidi2021review}. Similarly, RNNs excel at learning from sequential data, where past information influences future predictions, making them well-suited for text and time-series data. However, tabular data lacks such inherent spatial or temporal structures. In tabular formats, features do not follow any specific spatial or temporal order, and their relative positions carry no meaningful information. Reordering columns or rows does not alter the relationships between features, making models like CNNs and RNNs unsuitable without significant adaptation \cite{lecun1998gradient, lecun2015deep}. This absence of local correlations and temporal dependencies in tabular data makes it challenging for traditional deep-learning models to perform effectively, particularly when non-spatial relationships are critical. 

Recent research has sought to address these challenges by introducing novel architectures specifically designed to capture relational structures within tabular data. For example, models such as Dual-Route Structure-Adaptive Graph Networks (DRSA-Net) \cite{zheng2023deep} and Homological CNNs \cite{briola2023homological} employ graph-based and topologically constrained approaches to model the dependencies between features. Other approaches, like GOGGLE \cite{liu2023goggle}, focus on learning generative models by exploiting underlying relational structures, while TabularNet \cite{du2021tabularnet}  combines spatial and relational information using advanced techniques such as pooling and Graph Convolutional Networks (GCNs). These innovations represent significant progress in adapting deep learning architectures to the unique challenges posed by tabular data, paving the way for more effective modeling of complex, non-spatial relationships.

Similarly, Hellerstein, \cite{hellerstein2024learning} addresses the inherent challenges of working with tabular data, particularly when it lacks a grid-like structure typically seen in other data types such as images or text. The study focuses on automating the transformation of unstructured tables into tidy, relational forms suitable for analysis. It also introduces the idea that clean data tables can be considered as grids of cells, somewhat analogous to pixels in an image, where adjacent rows and columns may exhibit patterns. While deep learning models excel in pattern recognition in image grids, detecting such patterns in tabular data is far more difficult due to the diversity in how tables are structured and the absence of explicit spatial relationships. Ucar et al. \cite{ucar2021subtab} discussed the challenges posed by the lack of inherent spatial structure in tabular data. While image data benefits from spatial coherence (e.g., neighboring pixels are spatially correlated), and text or audio from semantic and temporal structures, tabular data lacks such clear patterns. This makes it difficult to apply common augmentation techniques like cropping or rotation, which are highly effective in domains such as image processing. To overcome these limitations, the authors propose the SubTab framework, which divides input features of tabular data into subsets, analogous to feature bagging or image cropping, to generate different views of the data. 

By reconstructing full data from these subsets, the framework forces the model to learn better representations of tabular data in a self-supervised setting, despite the absence of grid-like structure. This approach enables the model to discover patterns and relationships within tabular data that are not immediately apparent, and the results demonstrate that SubTab can achieve state-of-the-art performance on various datasets.

In an effort to refine this approach, Wang and Sun \cite{wang2022transtab} introduced TransTab, as shown in Figure 3 and Figure 4 below, a model that uses transformers to encode tabular data by treating rows (samples) and columns (features) as sequences. Figure 3 illustrates TransTab's ability to handle tasks like transfer learning, feature incremental learning, and zero-shot inference, demonstrating its adaptability across different tabular data tasks. Figure 4 details the framework, where categorical, binary, and numerical features are tokenized and processed through a gated transformer with multi-head attention, enabling efficient learning of feature interactions. This structured approach allows TransTab to handle variable-column tables and facilitates knowledge transfer, even across tables with different structures, enabling more effective learning and generalization across tasks and domains. The model focuses on learning generalizable representations, which can be applied to different datasets, overcoming the limitations imposed by the nonspatial nature of tabular data. The contextualization of columns and cells in TransTab introduces a structured way to interpret relationships within tabular data, enabling more effective learning and generalization across tasks and domains.

In a similar effort Ghorbani et al. \cite{ghorbani2021beyond} introduce the Feature Vectors method, which generates feature embeddings that capture both the importance and semantic relationships between features. Drawing inspiration from word embeddings in NLP, where words that frequently co-occur in the same context share similar embeddings, the authors apply a similar approach to features in tabular data. However, given that tabular data lacks natural co-occurrence structures, the authors propose using decision trees to extract co-occurrence relationships between features. By analyzing decision paths in tree-based models, they are able to create feature embeddings that preserve semantic relationships despite the nonspatial nature of tabular data. Also, Geisler and Binnig \cite{geisler2022introducing} discussed the challenges of applying existing model explanation methods, such as Local Interpretable Model-agnostic Explanations (LIME) and SHapley Additive exPlanations (SHAP), to tabular data. These methods, originally designed for data with spatial or temporal relationships, often fall short when applied to tabular data due to the absence of clear spatial or sequential patterns. To overcome this, they propose the Quest framework, which generates explanations in the form of relational queries specifically tailored for tabular data. Quest uses surrogate models and query-driven explanations to address the unique structure of tabular data, offering a more semantically rich and intuitive understanding of model behavior. By focusing on relational query predicates, Quest is able to explain not only why a model produced a particular output but also why it did not produce an alternative output. 

\begin{figure}[htp]
    \centering
    \includegraphics[width=0.98\linewidth]{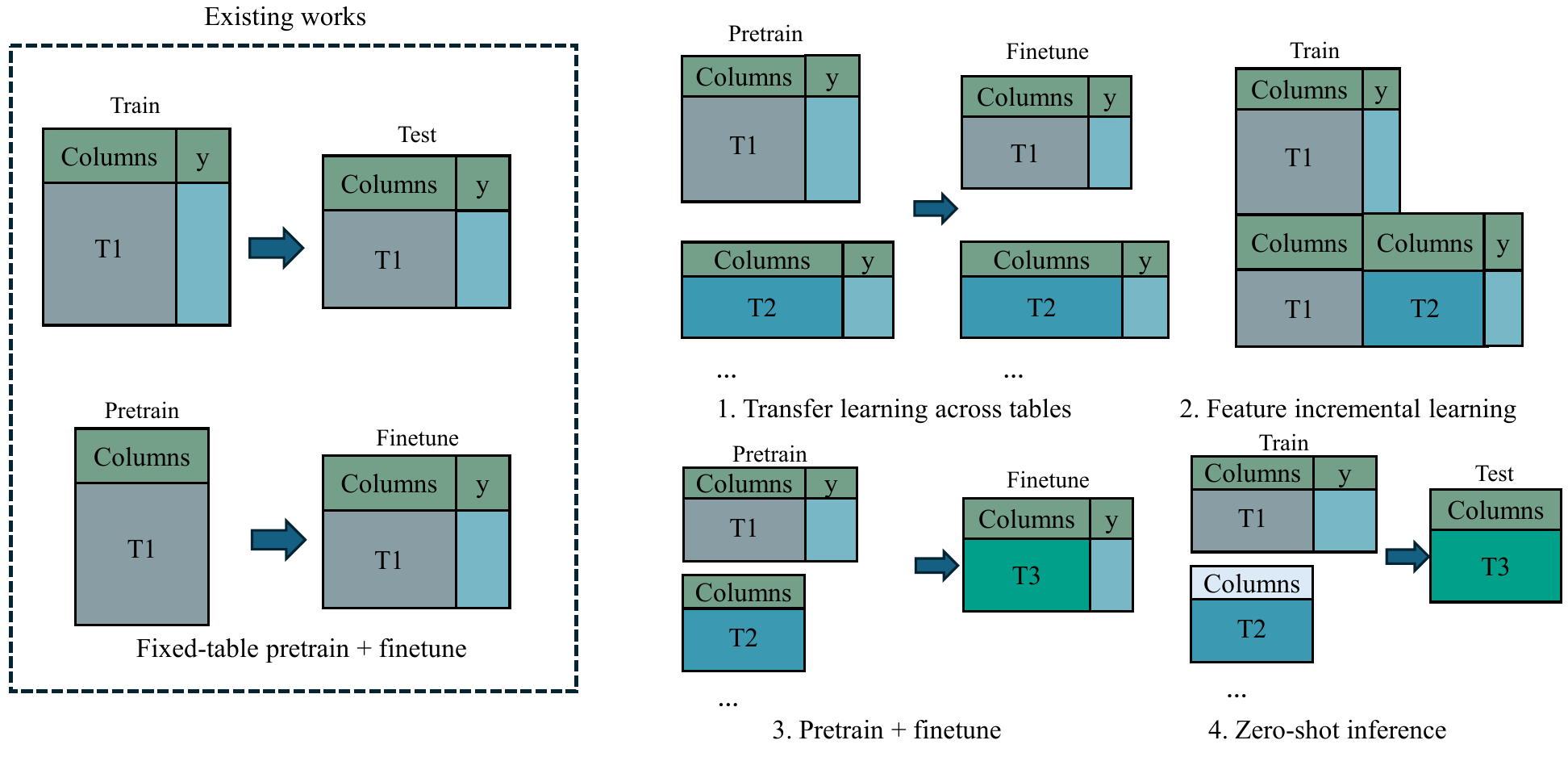}
    \caption{Demonstration of TransTab Tasks \cite{wang2022transtab}}
    \label{fig: TransTab Tasks}
\end{figure}

\begin{figure}[htp]
    \centering
    \includegraphics[width=0.98\linewidth]{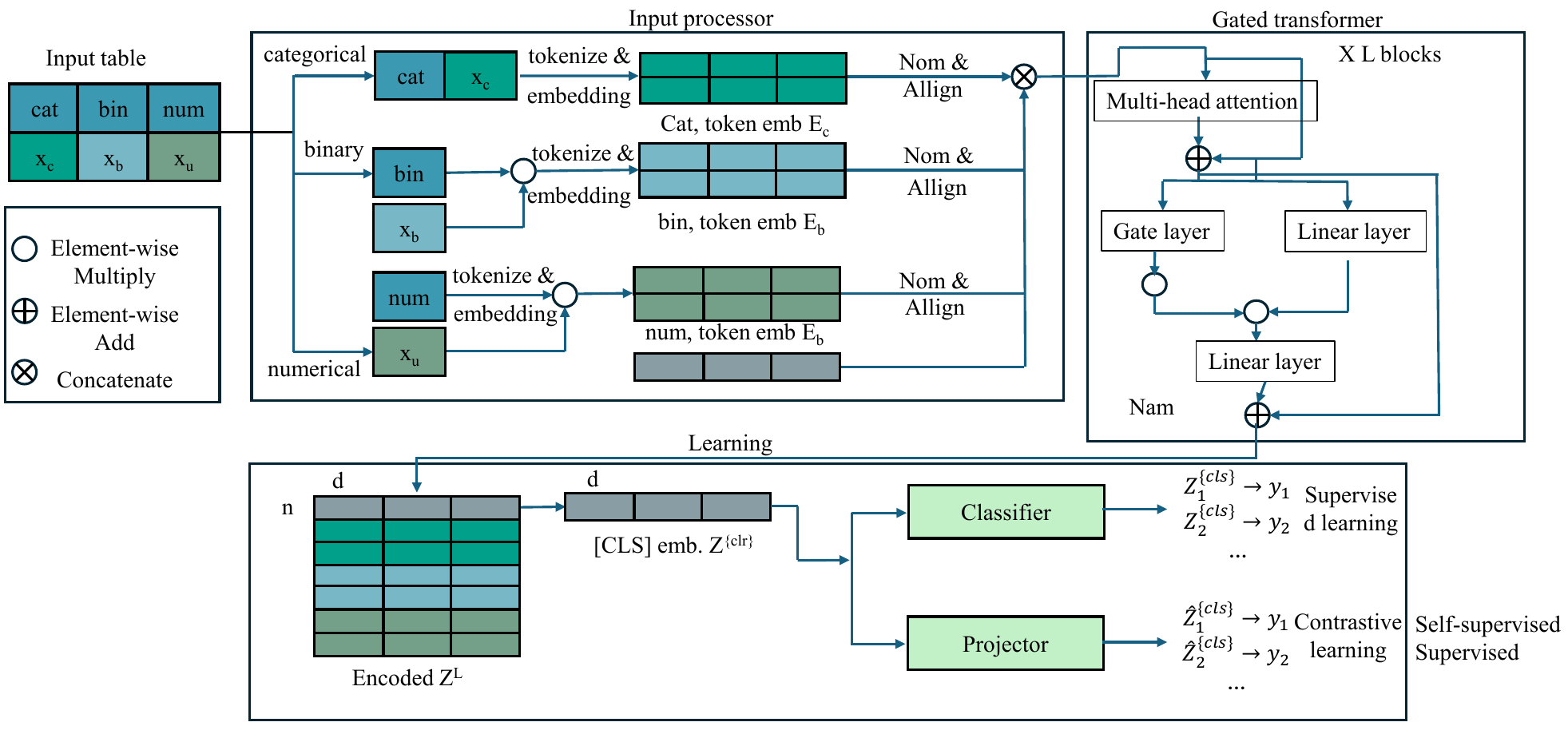}
    \caption{TransTab Framework \cite{wang2022transtab}}
    \label{fig:TransTab Framework}
\end{figure} 

Moving in the same direction Zhu et al. \cite{zhu2021converting} explain that CNNs excel when applied to data with spatial or temporal relationships, such as the arrangement of pixels in images or the sequential nature of text, allowing them to capture local patterns through convolutions. However, the absence of these structures in tabular data presents a significant challenge for CNN-based modeling. To address this challenge, the authors propose a novel algorithm called the Image Generator for Tabular Data (IGTD), which transforms tabular data into image-like structures by assigning tabular features to pixel positions while preserving feature relationships. This transformation introduces a form of spatial relationship in the data, allowing CNNs to process tabular data more effectively. The study demonstrates that these image representations help CNNs capture feature relationships and improve predictive performance compared to models trained on raw tabular data. The IGTD method addresses the lack of spatial or sequential dependencies in tabular data by creating artificial spatial relationships making it more compatible with deep learning models designed for structured data.

\subsection{Overfitting in Small Tabular Datasets}

Overfitting is a process where a model is unable to generalize and fits too closely to the training dataset. It can occur for a number of causes, including insufficient data samples and an inadequate training set to adequately represent all potential input data values. Overfitting is a significant challenge in deep learning, particularly when working with small datasets, as the model may end up memorizing the training data instead of learning generalizable patterns. To address this issue, various strategies have been proposed in the literature. One notable approach is transfer learning, where a model pre-trained on a large, related dataset is fine-tuned on a smaller target dataset. This helps to mitigate overfitting by leveraging the prior knowledge embedded in the pre-trained model, thus reducing the need for extensive data in the target domain \cite{jain2021deep, lecun2015deep}. Additionally, the use of techniques like image generation from tabular data and the application of CNNs are explored to handle small datasets more effectively. In their work, Koppe et al. \cite{koppe2021deep} highlight the importance of balancing the bias-variance trade-off in deep learning models trained on small datasets. They argue that overfitting occurs when models capture noise and peculiarities specific to the training data rather than generalizable patterns. To counter this, they recommend regularization techniques such as dropout and the incorporation of domain knowledge during model training. These methods help to constrain the model's flexibility and reduce the likelihood of overfitting. LeCun et al. \cite{lecun2015deep} pointed out that while deep networks can learn complex representations, their flexibility can lead to memorization of the training data. To mitigate this, they suggest using unsupervised pre-training and data augmentation, which have proven effective in improving the generalization capabilities of deep learning models.

While deep learning has made significant strides in domains like computer vision and NLP, its application to tabular data has proven more difficult. This disparity can be attributed to the fundamental differences in data volume and structure between these domains. In computer vision and NLP, large-scale datasets such as ImageNet (with millions of labeled images) and GPT-3's massive corpora help models learn complex representations without overfitting. By contrast, tabular datasets commonly found in fields like engineering, healthcare, and finance are often significantly smaller, comprising only hundreds or thousands of samples. This size limitation makes it challenging to train deep learning models effectively and increases the risk of overfitting due to insufficient data diversity. As explained in few studies \cite{zhao2017research, brigato2021close} outlined, deep learning models excel in image classification due to their ability to learn from vast amounts of data, but struggle when applied to small datasets, leading to overfitting. The high number of parameters in deep models, such as those found in CNNs, further exacerbates the problem when there are limited training examples. Similarly, Jain et al., \cite{jain2021deep} emphasize that deep learning models tend to perform poorly on tabular data due to the dataset’s heterogeneous nature. Deep models, which rely on large-scale homogeneous data (as seen in computer vision and NLP), often fail to generalize well in domains where tabular data is used, causing overfitting after only a few epochs.

\begin{figure}[htp]
    \centering
    \includegraphics[width=0.98\linewidth]{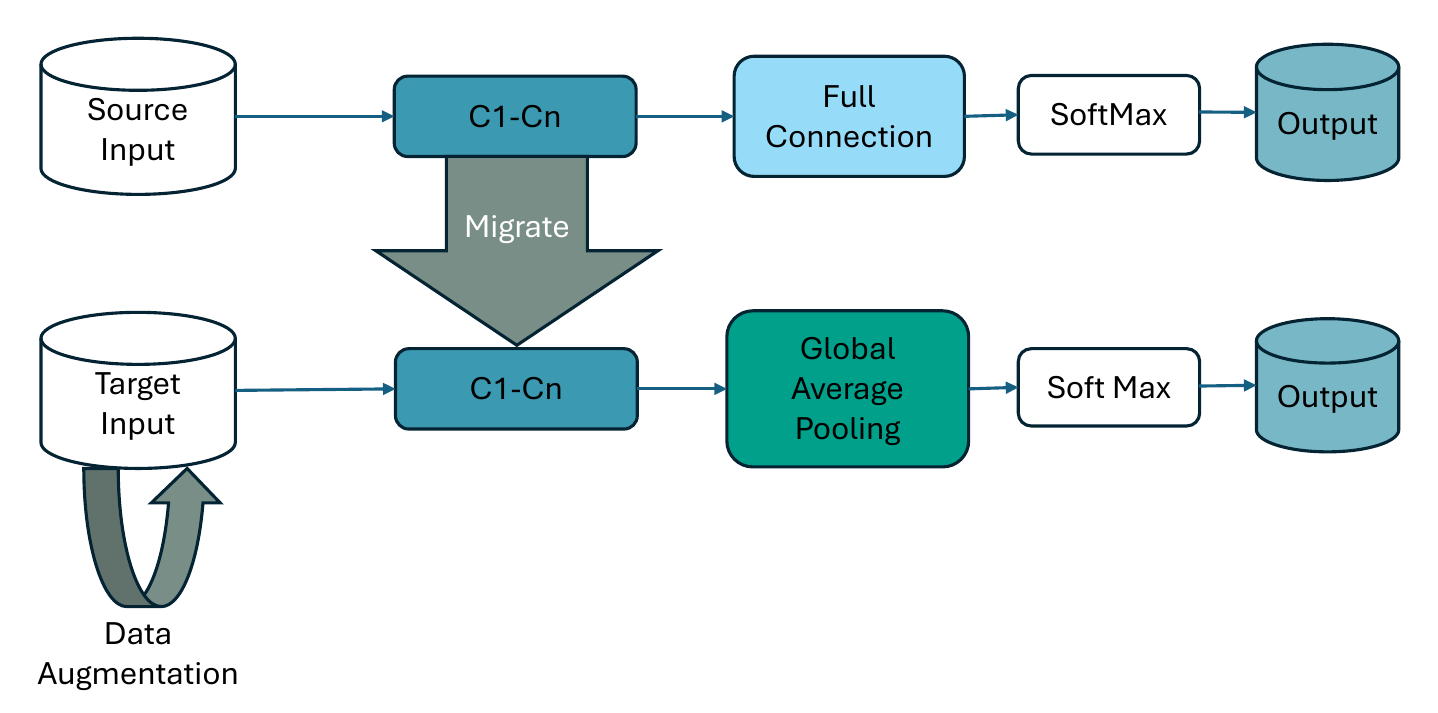}
    \caption{Transfer Learning in Pre-Trained-TLCNN \cite{zhao2017research}}
    \label{fig:Pre-Trained-TLCNN}
\end{figure} 

To overcome this, researchers have proposed several methods to address the challenges of training deep learning models on small tabular datasets. One prominent approach is transfer learning, which has been successful in mitigating overfitting in small datasets for computer vision and NLP. For example, Zhao \cite{zhao2017research} proposes using transfer learning combined with data augmentation to tackle overfitting in small datasets as shown in Figure 5. By pre-training a CNN on large datasets like ImageNet, and then fine-tuning it on smaller datasets, the model can leverage previously learned representations to improve performance on the target task. Jain et al., \cite{jain2021deep} extend this concept to tabular data by converting tabular datasets into image representations using techniques like IGTD and SuperTML. These methods allow deep learning models, particularly CNNs, to be applied to tabular data by transforming it into an image-like format that can take advantage of pre-trained models, thus reducing overfitting. Horenko \cite{horenko2020scalable} introduces the entropy-optimal scalable probabilistic approximations algorithm algorithm to breach the overfitting barrier at a fraction of the computational cost. Badger \cite{badger2022smalllanguagemodelstabular} demonstrates the effectiveness of small language models in processing tabular data without extensive preprocessing, achieving record classification accuracy. Another promising solution is the HyperTab method proposed by Wydmański et al. \cite{wydmanski2023hypertab}, which uses a hypernetwork-based approach to build an ensemble of neural networks specialized for small tabular datasets. A general HyperTab structure is shown in Figure 6. By employing feature subsetting as a form of data augmentation, HyperTab virtually increases the number of training samples without altering the number of parameters. This approach allows the model to generalize better by preventing overfitting, particularly on small datasets.
\begin{figure}[htp]
    \centering
    \includegraphics[width=0.98\linewidth]{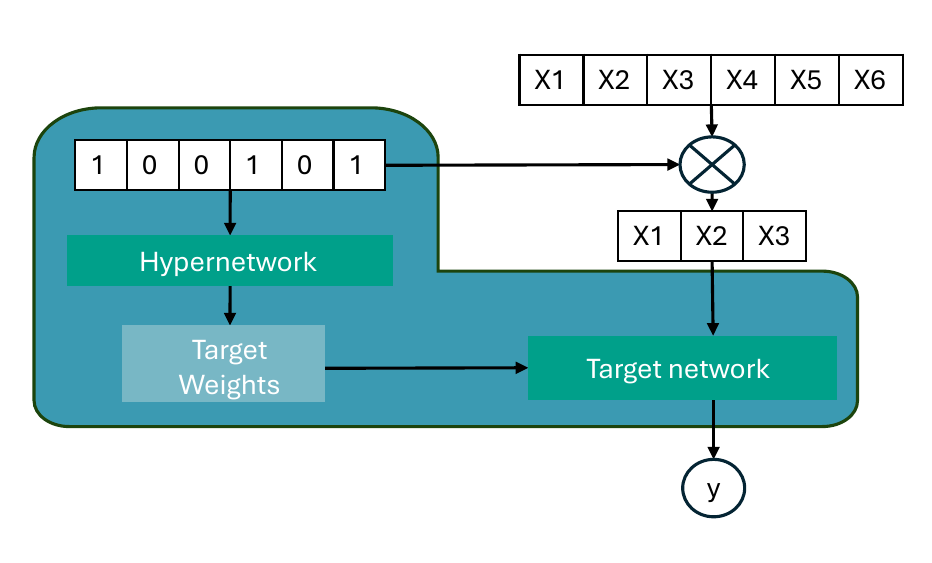}
    \caption{General HyperTab Structure \cite{wydmanski2023hypertab}}
    \label{fig:HyperTab}
\end{figure}

\section{Historical Evolution of Tabular Deep Learning}
\subsection{Classical Approaches}
Before the emergence of deep learning, traditional machine learning models like  Support Vector Machines  (SVMs), linear and logistic regression, and tree-based methods have long been the preferred choice for tabular data analysis \cite{singh2023embeddingstabulardatasurvey}. These classical approaches were well-suited for small-scale tabular datasets but limited to classification and regression tasks. These models were not only highly interpretable, allowing users to understand and explain predictions, but they also handled both numerical and categorical data well. They were ideally suited for small to medium-sized datasets, as they required less computational power and were quick to train. Despite the rise of deep learning, these traditional models are still preferred in certain cases today. Additionally, their faster training and deployment times make them ideal for applications where real-time decisions are necessary. These classical approaches were well-suited for small-scale tabular datasets but limited to classification and regression tasks. 
However, these traditional models are not without their limitations. For instance, Clark et al. \cite{clark2023dealing} point out that logistic regression models can encounter problems such as complete and quasi-complete separation, where the model either perfectly or nearly perfectly separates the data. This can lead to extremely large or infinite coefficient estimates, making statistical inference unreliable. Moreover, logistic regression is particularly sensitive to small sample sizes, especially when dealing with low-frequency categorical variables, which can worsen separation issues. To counter this, covariates are often removed, or categories merged, but such actions can lead to oversimplification and a reduction in the model's predictive power. Similarly, Carreras et al. \cite{valero2023comparing} highlight several drawbacks of SVMs, particularly in the soft margin variant. These include the risk of overfitting, increased computational complexity when feature selection is involved, and the non-convex nature of the optimization problem, which complicates finding optimal solutions. Additionally, achieving a complete Pareto front in multi-objective optimization is difficult due to new parameters leading to similar solutions despite weight variations.

Expanding on the strengths and limitations of traditional models like SVMs and logistic regression, models like Decision Trees, Naive Bayes, and early Neural Networks also relied on manual feature engineering, requiring domain expertise to select relevant features \cite{inproceedings}. This process, while labor-intensive, allowed these models to perform effectively on smaller datasets. Abrar and Samad \cite{abrar2023deep} emphasize that while fully connected deep neural networks have become popular in recent years, traditional machine learning models like gradient boosting trees still outperform deep learning models in many cases, particularly when dealing with tabular data containing uncorrelated variables. This study highlights that traditional models like gradient boosting trees are superior when deep models fail, especially in datasets that lack the strong correlations often found in real-world data. Additionally, these classical models do not face the overfitting challenges or high computational costs associated with deep learning. Unlike deep learning models, which tend to overly smooth the relationships in data, tree-based methods can accurately partition the feature space and learn locally constant functions, making them ideal for datasets with irregular target functions. These models are also more robust to uninformative features, which are common in tabular datasets, while neural networks, especially multi-layer perceptrons (MLPs), struggle with irrelevant or redundant features, negatively impacting their performance \cite{grinsztajn2022treebasedmodelsoutperformdeep}. Additionally, tree-based models preserve the original orientation of the data, which is important in tabular datasets where features often carry individual meanings, such as age or income. Tree-based models excel at handling the complexities of tabular data by partitioning the feature space in a way that captures non-linear relationships, which deep learning models often struggle to do without overfitting.  A study by Fayaz et al. \cite{Fayaz2022} found that even when applied to large datasets, traditional models like XGBoost consistently outperformed state-of-the-art deep learning models, especially when the data lacked strong correlations that deep learning models rely on.

While tree-based models offer numerous advantages for tabular data, they are not without their challenges. Tree-based models, such as decision trees, face several challenges when dealing with tabular data. One key issue is scalability, particularly with large datasets \cite{schidler2024sat}. As the complexity of the dataset increases, decision trees tend to grow deep, which significantly raises runtime and computational costs. This scalability problem is especially pronounced in models designed to handle large datasets, as they struggle to balance depth and size without sacrificing accuracy. Another major drawback is overfitting, where deep trees tend to memorize the training data, including noise and irrelevant features, leading to poor generalization on unseen data \cite{costa2023recent}. Although techniques like pruning can help mitigate this, they may reduce model accuracy. Traditional decision trees also rely on univariate splits, which can oversimplify complex relationships between features in tabular data, often leading to unnecessarily large trees. While multivariate trees can capture more intricate patterns, they come with added complexity and reduced interpretability. Additionally, decision trees often struggle with imbalanced data, as they tend to be biased towards the majority class \cite{marudi2024decision}. Techniques like SMOTE or cost-sensitive learning are required to address this, but these methods increase computational overhead. Furthermore, decision trees that use models in their leaves, known as model trees, face significant increases in training time and complexity, especially when evaluating a large number of candidate splits across a wide range of features. As tabular deep learning transitioned from classical methods, foundational models emerged that addressed many of the limitations of tree-based approaches. Table 1 outlines these key models, showcasing their core architectures and training methodologies, which laid the groundwork for the more advanced techniques seen in modern tabular deep learning.

\thispagestyle{empty}
\begingroup
\fontsize{8pt}{8pt}\selectfont
\begin{longtable}{p{0.14\linewidth} p{0.2\linewidth} p{0.18\linewidth} p{0.3\linewidth}}
\caption{Timeline of DL Models for Tabular Data (2016-20)}
\label{tab:timeline_dl_models_2016_20} \\
\toprule
\textbf{\makecell{Model \\ (Year) \\ Source}} & 
\textbf{Architecture} & 
\textbf{\makecell{Training \\ Efficiency}} & 
\textbf{Main Features} \\
\midrule
\endfirsthead
\toprule
\textbf{\makecell{Model \\ (Year) \\ Source}} & 
\textbf{Architecture} & 
\textbf{\makecell{Training \\ Efficiency}} & 
\textbf{Main Features} \\
\midrule
\endhead
\midrule
\endfoot
\bottomrule
\endlastfoot
VIME (2020) \cite{yoon2020vime} & Neural network + masked and feature-vector estimation & Self and semi-supervised learning; Moderate & Self-supervised learning and contextual embedding \\
\midrule
NON (2020) \cite{luo2020network} & Field-wise + across field + operation fusion network & Supervised learning; Moderate & Network-on-network model \\
\midrule
Net-DNF (2020) \cite{katzir2020net} & Affine literals + conjunction + output layer & Supervised learning; Moderate & Structure-based on disjunctive normal form \\
\midrule
TabTransformer (2020) \cite{huang2020tabtransformer} & Transformer-based architecture + contextual embeddings & Supervised learning, semi-supervised learning; Moderate & Transformer network for categorical data \\
\midrule
TabNet (2019) \cite{arik2021tabnet} & Sequential Attention + Sparse Feature Selection + Feature Transformer & Supervised learning; Moderate & Sequential attention structure \\
\midrule
NODE (2019) \cite{popov2019neural} & Differentiable oblivious decision trees (ODT) & Supervised learning; Moderate & Differentiable decisions are made with classic decision trees via the entmax transformation \\
\midrule
DeepGBM (2019) \cite{ke2019deepgbm} & Hybrid approach integrating GBDT with NN & Supervised learning; Moderate to high & Two DNNs distill knowledge from decision tree \\
\midrule
SuperTML (2019) \cite{sun2019supertml} & CNN-based & Supervised learning; Moderate & Transform tabular data into images for CNNs \\
\midrule
xDeepFM (2018) \cite{lian2018xdeepfm} & Hybrid neural network & Supervised learning; Moderate to high & Embedding Layer + compressed interaction network + DNN \\
\midrule
TabNN (2018) \cite{ke2018tabnn} & Automatic feature grouping + recursive encoder with shared embedding & Supervised learning; Moderate to high & DNNs based on feature groups distilled from GBDT \\
\midrule
RLN (2018) \cite{shavitt2018regularization} & Regularization mechanism + sparse network & Supervised learning; Moderate & Hyperparameters regularization scheme \\
\midrule
DeepFM (2017) \cite{guo2017deepfm} & Factorization machines + deep neural networks & Supervised learning; Moderate to high & Combining low- and high-order feature interactions + shared embedding layer \\
\midrule
Wide\&Deep (2016) \cite{cheng2016wide} & Memorization (wide component) and generalization (deep component) & Supervised learning; High & Cross-product feature transformations for memorization + Embedding layer for categorical features \\
\end{longtable}
\endgroup

\subsection{Shallow Neural Networks}
Recent research highlights the ongoing challenges of applying deep learning to tabular data. Despite deep learning's success in image and text domains, tree-based models like XGBoost and Random Forests continue to outperform neural networks on medium-sized tabular datasets \cite{grinsztajn2022tree}. This performance gap persists even after extensive hyperparameter tuning. Researchers have identified key challenges for developing tabular-specific neural networks, including robustness to uninformative features, preserving data orientation, and learning irregular functions \cite{borisov2022deep, shwartzziv2021tabulardatadeeplearning}.

Katzir et al. \cite{katzir2020net} introduced the Net-DNF architecture, embedding inductive biases similar to gradient-boosting decision trees (GBDTs), to address the shortcomings of FCNs in tabular data tasks. Their experiments demonstrated that Net-DNF outperformed traditional FCNs, particularly on large-scale tabular datasets, underscoring the limitations of conventional neural architectures for this domain. Similarly, Borisov et al. \cite{borisov2022deep} critiqued deep neural networks for tabular data, noting that early attempts with shallow and FCN often failed to match the performance of tree-based models like decision trees and GBDTs. They emphasized that FCNs struggle with the unique challenges of tabular data, such as handling categorical variables, missing entries, and imbalanced datasets, and that feature engineering alone rarely closes the performance gap. In line with this, Abutbul et al. \cite{abutbul2020dnf} proposed DNF-Net, a neural architecture incorporating Boolean logic and feature selection, which consistently outperformed FCNs on large-scale tabular classification tasks. Chauhan and Singh \cite{chauhan2018review} and Abrar and Samad \cite{abrar2022perturbation} similarly recognized the use of shallow networks, such as MLPs, for tabular data but highlighted their limitations, including overfitting and limited research focus compared to more advanced methods. While MLPs are effective, they are often surpassed by specialized or more sophisticated architectures in handling the complexities of tabular data.

Early applications of shallow neural networks, particularly FCNs, to tabular data often underperformed compared to specialized models like GBDTs. However, recent studies have shown that with proper tuning and architectural enhancements, neural networks can rival or surpass GBDTs. Chen et al. \cite{chen2024excelformer} emphasized the efficiency of shallow networks in handling unordered tabular data, while Erichson et al. \cite{erichson2020shallow} demonstrated their competitiveness in tasks like fluid dynamics, where fast training and regularization were key advantages. Rubachev et al. \cite{rubachev2022revisiting} further noted that with optimized tuning and techniques like unsupervised pre-training, shallow networks can close the performance gap with GBDTs, although this gain is context-dependent. Fiedler \cite{fiedler2021simple} introduced structural innovations such as Leaky Gates and Ghost Batch Norm, which significantly enhanced MLPs for tabular data, enabling them to outperform GBDTs in several cases. Figure 7 shows the original and the modified MLP+ model. These advancements demonstrate that shallow networks when effectively optimized, can meet the unique challenges of tabular data and compete with traditional models. 

\begin{figure}[htp]
    \centering
    \includegraphics[width=0.98\linewidth]{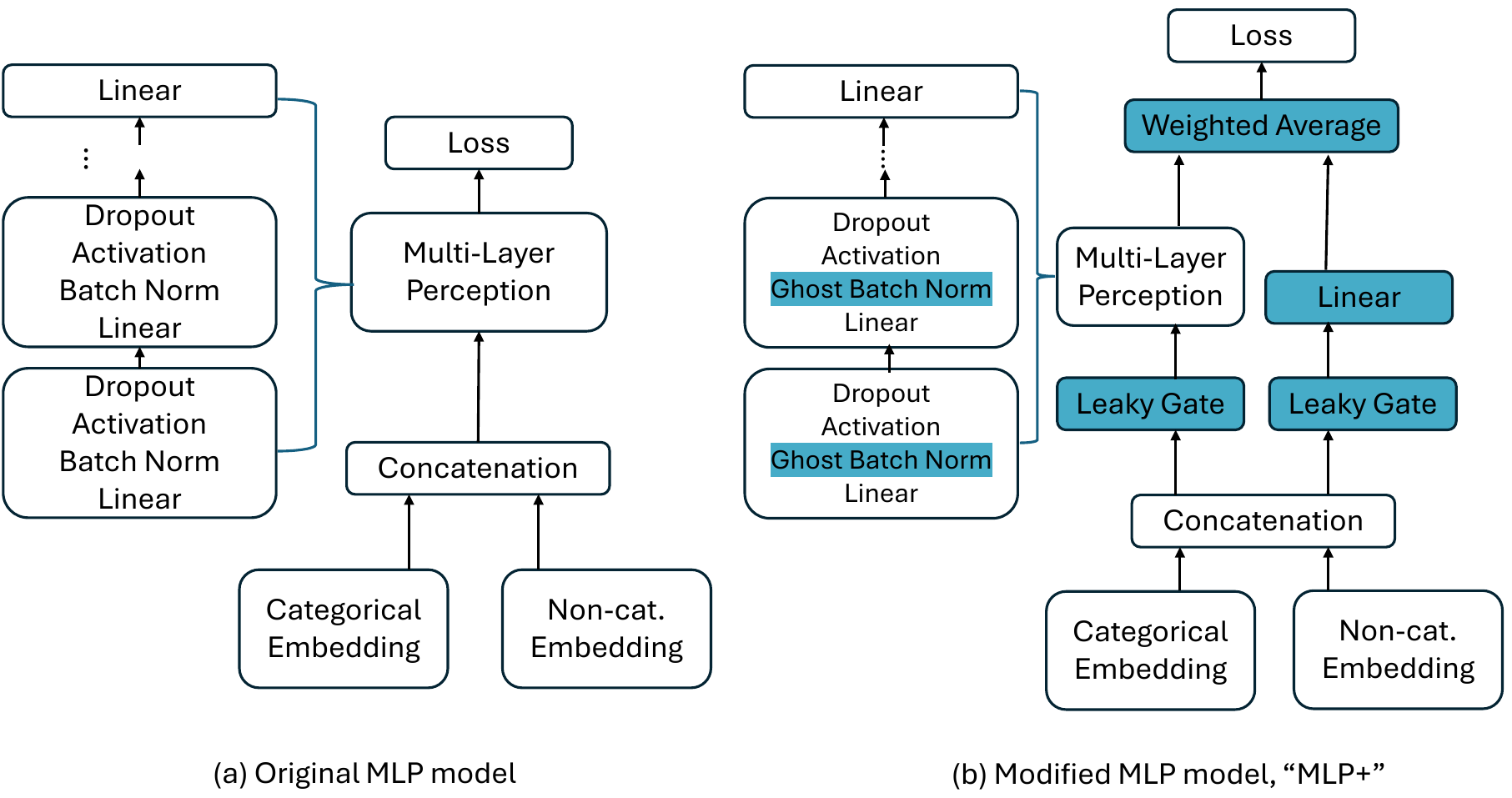}
    \caption{Original and Modified MLP Models \cite{fiedler2021simple}}
    \label{MLP}
\end{figure} 

These findings align with the broader consensus that the standard architecture of FCNs often lacks the inductive biases necessary for effectively modeling the complexities of tabular data, such as categorical variables, missing data, and imbalanced datasets. Specialized neural networks are frequently required to address these challenges. However, Grinsztajn et al. \cite{grinsztajn2022tree} offer a more optimistic view, demonstrating that shallow FCNs, like MLPs, can remain competitive when combined with regularization techniques to mitigate overfitting and generalization issues. They further suggest that even simple architectures, such as ResNet, can match the performance of more advanced models, indicating that, with proper modifications, shallow networks can still play a valuable role in handling tabular data. 

\subsection{Initial Breakthroughs}
TabNet and NODE represent significant advancements in the application of deep learning to tabular data, addressing longstanding challenges in performance, interpretability, and efficiency. This research explores how these models tackle issues inherent to tabular data, such as handling heterogeneous features and preventing overfitting while introducing innovations that set them apart from classical machine learning methods and earlier neural network approaches.

\subsubsection{\textbf{TabNet}}
TabNet is a deep learning architecture specifically designed to address the challenges associated with applying neural networks to tabular data. Unlike image or text data, tabular data often consists of heterogeneous features, making it difficult for traditional deep learning models, such as MLPs, to efficiently capture the relationships between the features. Classical machine learning models have traditionally excelled in this domain due to their ability to handle the complex decision boundaries of tabular data. However, deep learning offers potential advantages, such as end-to-end learning and the ability to integrate with other data types, which TabNet leverages through its novel architecture. TabNet introduces several key innovations to overcome these challenges. One of the central features of TabNet is its sequential attention mechanism, which allows the model to select the most important features for each decision step dynamically \cite{arik2021tabnet}. This instance-wise feature selection sets TabNet apart from other models, as it can tailor the features used for each individual input rather than relying on a fixed set of features for all instances. This dynamic feature selection leads to more efficient learning by focusing the model’s capacity on the most relevant features, which is particularly beneficial for tabular data that may have a mix of irrelevant or redundant information. Figure 8 below shows the architecture of the encoder and decoder of a TabNet model.

\begin{figure}[htp]
    \centering
    \includegraphics[width=0.98\linewidth]{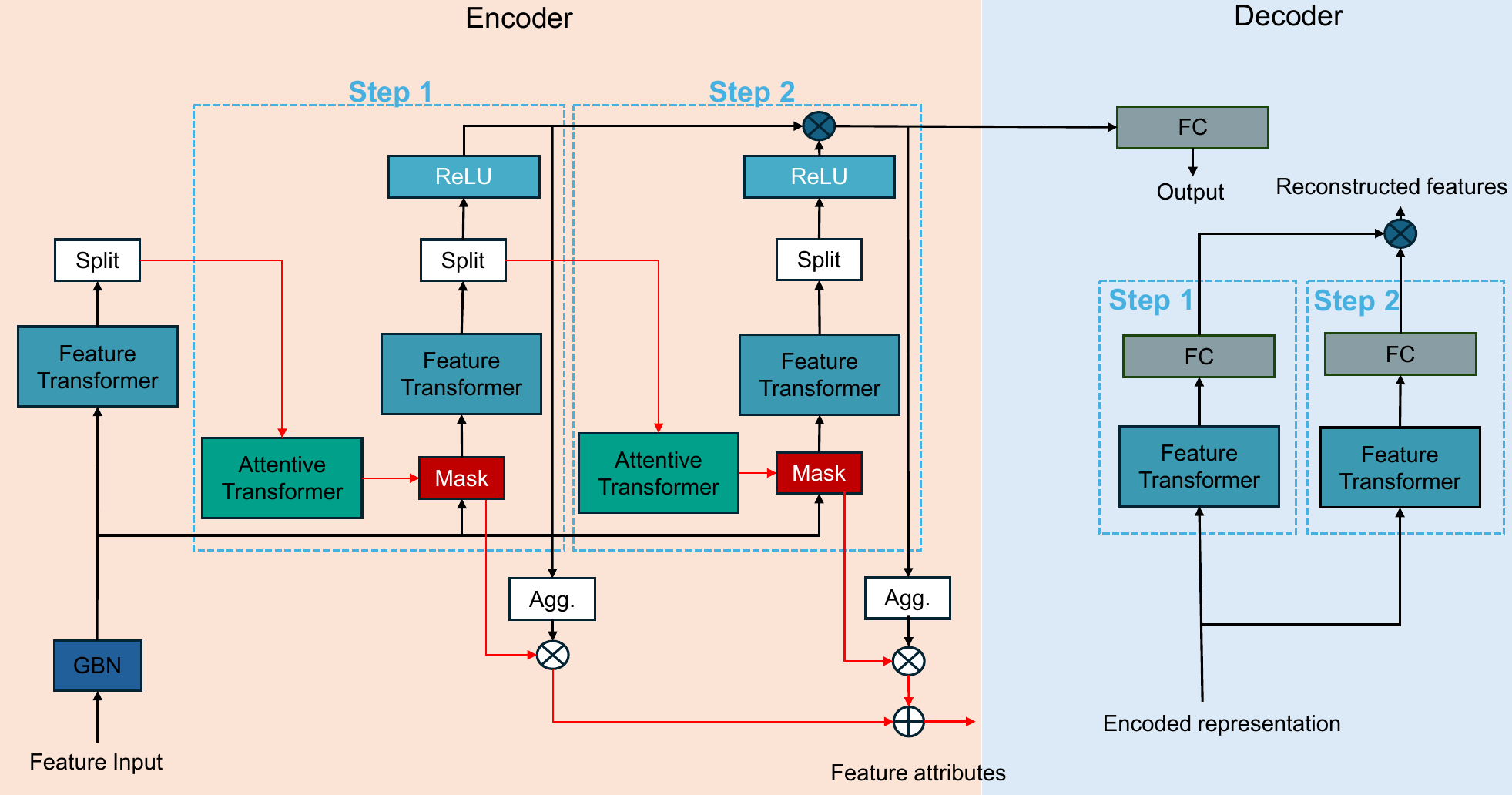}
    \caption{TabNet Encoder and Decoder Architecture \cite{wa2023stable}}
    \label{fig:TabNet}
\end{figure} 

TabNet offers significant advancements in interpretability compared to both classical machine learning models and traditional neural networks. By integrating sparse attention and feature masks, TabNet enhances performance while providing insights into which features influence predictions. This results in both local and global interpretability, making it possible to visualize individual feature importance and quantify overall contributions \cite{ta2023tabnet}. The newer InterpreTabNet builds on this by improving feature attribution methods, further enhancing interpretability, and making the model's decisions more transparent at both local and global levels \cite{wa2023stable}.

TabNet excels in working with raw tabular data without the need for extensive preprocessing or manual feature engineering, which is often required by traditional models like GBDTs. Its end-to-end learning capabilities allow TabNet to directly process raw data, simplifying workflows while maintaining high performance. Additionally, TabNet introduces self-supervised learning, a novel feature for tabular data, where it can pre-train on unlabeled data using masked feature prediction to improve performance on supervised tasks, particularly when labeled data is scarce. Evaluated on various datasets, TabNet has been shown to outperform or match state-of-the-art models, including GBDTs, in both classification and regression tasks. For example, in facies classification, it achieves superior accuracy compared to traditional tree-based models and other deep learning architectures like 1D-CNNs and MLPs \cite{ta2023tabnet}. Its flexible architecture, which incorporates sequential feature transformers and attention mechanisms, enhances generalization across different domains, while the use of sparse attention ensures interpretability, addressing a key limitation of traditional deep learning models.

TabNet, despite its innovations like interpretability and sparse attention, is often outperformed by XGBoost across various datasets, requiring more hyperparameter tuning and showing less consistent results \cite{fayaz2022deep}. Additionally, TabNet's training time is significantly longer, making it less practical for quick iterations or real-time applications \cite{lewandowska2022xgboost}. It is also prone to overfitting on smaller datasets due to its complex architecture, especially when not tuned correctly.

\subsubsection{\textbf{Neural Oblivious Decision Ensembles (NODE)}}
NODE have been proposed to address the specific challenges of applying deep learning to tabular data, which has traditionally been dominated by tree-based models like GBDT. Popov et al. (2019) identified the key limitations of deep learning models in handling tabular data, primarily due to their inability to outperform GBDTs consistently. To bridge this gap, NODE was introduced as a deep learning architecture that generalizes ensembles of oblivious decision trees, offering end-to-end gradient-based optimization and multi-layer hierarchical representation learning. This design allows NODE to capture complex feature interactions within tabular data, a task where traditional deep learning models often fall short. One of NODE's key innovations is the use of differentiable oblivious decision trees, where splitting decisions are made through the entmax transformation, allowing for soft, gradient-based feature selection. This approach makes the decision-making process more flexible and differentiable, unlike conventional decision trees that rely on hard splits.
   
Additionally, NODE's multi-layer architecture is designed to capture both shallow and deep interactions within tabular data, effectively functioning as a deep, fully differentiable GBDT model trained end-to-end via backpropagation \cite{popov2019neural}. The architecture of NODE stacks multiple layers of differentiable oblivious decision trees, which enables NODE to outperform existing tree-based models in many tasks. Furthermore, NODE enhances computational efficiency by allowing pre-computation of feature selectors, significantly speeding up inference without sacrificing accuracy. Joseph \cite{joseph2021pytorchtabularframeworkdeep} explored NODE within the PyTorch Tabular framework, which simplifies deep learning for tabular data by offering a unified API that integrates both NODE and TabNet. This framework addresses the complexity of training deep learning models compared to traditional machine learning libraries like Scikit-learn, making advanced models more accessible for practitioners and researchers. Fayaz et al. \cite{fayaz2022deep} compared NODE, TabNet, and XGBoost, noting that while NODE introduces key innovations such as handling mixed data types and data imbalance, it often requires more hyperparameter tuning than XGBoost. However, combining NODE with XGBoost enhances performance, showing NODE's strength in complementing traditional models for tabular data.

\section{Recent Advances in Tabular Deep Learning}
While a previous study \cite{borisov2022deep} provides a structured overview of deep learning for tabular data, focusing on challenges like handling categorical variables, data transformation, and model comparison, this survey takes a different approach by emphasizing the historical evolution and algorithmic advancements in the field. We highlight the development of more recent models such as MambaNet, SwitchTab, and TP-BERTa, showing how these architectures have evolved to address the unique complexities of tabular data. By exploring advancements in attention mechanisms, hybrid architectures, and other recent breakthroughs, this survey underscores the transformation of deep learning models into more efficient, scalable, and interpretable solutions. Unlike previous work, this study does not focus on model comparison, as a comprehensive evaluation across models requires a separate analysis tailored to various types of tabular data.

In the rapidly evolving field of tabular deep learning, significant improvements have been made with each year bringing new architectures designed to address the increasing complexity of tabular data. Recent models, such as HyperTab and GANDALF, push the boundaries of scalability and interpretability, offering enhanced methods for handling heterogeneous features and high-dimensional data. These newer architectures build upon foundational work, leading to marked performance improvements over traditional approaches. As shown in Figure 9, the evolution of tabular deep learning highlights key contributions, ranging from Semek et al. \cite{samek2019explainable} and Arik et al. \cite{arik2021tabnet} in 2019 to the most recent developments, arranged by citation count to showcase the growing impact of this research. 

 \begin{figure}[htp]
    \centering
    \includegraphics[width=0.98\linewidth]{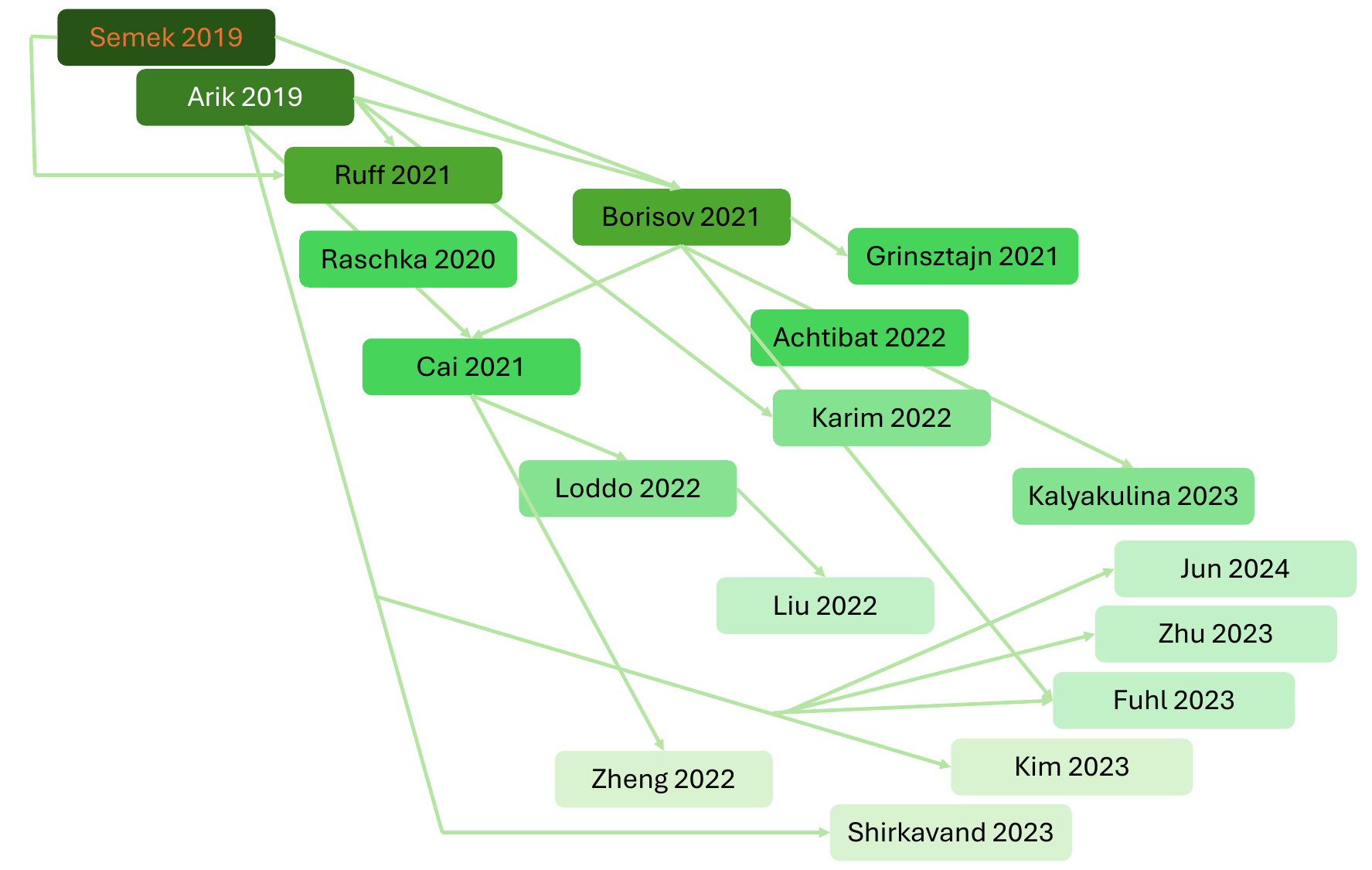}
    \caption{A Timeline of Tabular Deep Learning Papers}
    \label{fig:Timeline}
\end{figure}

Building on these developments, Table 2 presents a timeline of major models introduced during this period, detailing their architectures and key performance traits. These models highlight the significant breakthroughs in tabular deep learning, from hybrid architectures to advanced attention mechanisms, which have propelled performance and scalability forward.

\thispagestyle{empty}
\begingroup
\fontsize{8pt}{8pt}\selectfont
\begin{longtable}{p{0.12\linewidth} p{0.2\linewidth} p{0.18\linewidth} p{0.3\linewidth}}
\caption{Timeline of DL Models for Tabular Data (2021-22)}
\label{tab:timeline_dl_models_2021_22} \\
\toprule
\textbf{\makecell{Model \\ (Year) \\ Source}} & 
\textbf{Architecture} & 
\textbf{\makecell{Training \\ Efficiency}} & 
\textbf{Main Features} \\
\midrule
\endfirsthead
\toprule
\textbf{\makecell{Model \\ (Year) \\ Source}} & 
\textbf{Architecture} & 
\textbf{\makecell{Training \\ Efficiency}} & 
\textbf{Main Features} \\
\midrule
\endhead
\midrule
\endfoot
\bottomrule
\endlastfoot
TabLLM (2022) \cite{hegselmann2023tabllm} & Large language models & Few-shot supervised learning; Moderate & Serializes tabular data into natural language strings \\
\midrule
TabDDPM (2022) \cite{kotelnikov2023tabddpm} & Multinomial diffusion + gaussian diffusion & Supervised learning; Moderate to high & Multinomial and gaussian diffusion to handle categorical and numerical features \\
\midrule
Ptab (2022) \cite{liu2022ptab} & Pre-trained language model architecture & Supervised and self-supervised learning; N/A & Uses three-stage training strategy (modality transformation, masked-language fine-tuning, and classification fine-tuning) \\
\midrule
GANDALF (2022) \cite{joseph2022gate} & Gated feature learning unit (GFLUs) & Supervised learning; High & Uses GFLUs with learnable feature masks and hierarchical gating mechanisms \\
\midrule
ARM-Net (2021) \cite{cai2021arm} & Exponential neurons + gated attention mechanism + sparse softmax & Supervised learning; High & Adaptive relational modeling with multi-head gated attention network \\
\midrule
NPT (2021) \cite{kossen2021self} & Attention-based NN + datapoints + attributes & Self-supervised learning; Moderate to Low & Process the entire dataset at once, use attention between data points \\
\midrule
SAINT (2021) \cite{somepalli2021saint} & Hybrid architecture with both self-attention and intersample attention mechanisms & Self-supervised contrastive learning + supervised learning; Moderate & Attention over both rows and columns \\
\midrule
Regularized DNNs (2021) \cite{kadra2021well} & Plain Multilayer perceptron & Supervised learning; Moderate to high & A “cocktail” of regularization techniques \\
\midrule
Boost-GNN (2021) \cite{ivanov2021boost} & GBDT + GNN & Semi-supervised learning; Moderate to high & GNN on top decision trees from the GBDT algorithm \\
\midrule
DNN2LR (2021) \cite{liu2020dnn2lr} & DNN + LR & Supervised learning; High & Calculate cross-feature fields with DNNs for LR \\
\midrule
IGTD (2021) \cite{zhu2021converting} & CNN-based neural network & Supervised learning; High & Transform tabular data into images for CNNs \\
\midrule
SCARF (2021) \cite{bahri2021scarf} & Encoder + pre-train head network & Self-supervised contrastive, semi-supervised, fully supervised learning; N/A & Self-supervised contrastive learning, random feature corruption \\
\end{longtable}
\endgroup

\subsection{TabTransformer}
The TabTransformer model introduces significant advancements in tabular deep learning by leveraging attention mechanisms and hybrid architectures to address the unique challenges posed by tabular data \cite{huang2020tabtransformertabulardatamodeling}. At its core, TabTransformer employs multi-head self-attention layers adapted from the Transformer architecture, traditionally used in NLP, to capture complex feature interactions and dependencies across the dataset as seen in Figure 10. This attention mechanism enables the model to effectively capture relationships between features, making it particularly useful for datasets with numerous categorical variables.

\begin{figure}[htp]
    \centering
    \includegraphics[width=0.98\linewidth]{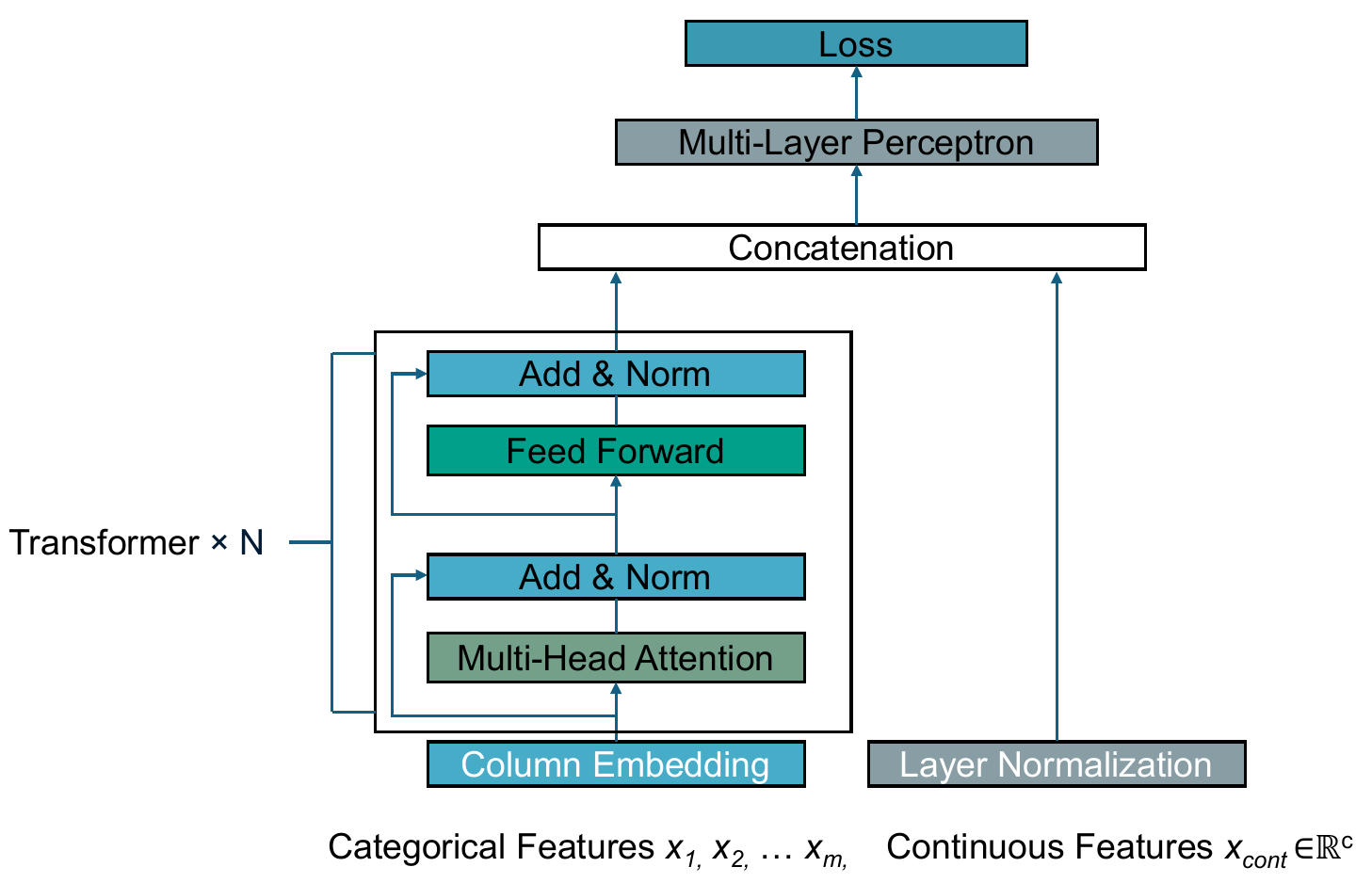}
    \caption{TabTransformer Architecture \cite{huang2020tabtransformertabulardatamodeling}}
    \label{fig:TabTransformer}
\end{figure} 

The TabTransformer architecture combines transformer layers with MLP components, forming a hybrid structure optimized for tabular data. Categorical features are embedded using a column embedding layer, which transforms each category into a dense, learnable representation. These embeddings are passed through Transformer layers, which aggregate contextual information from other features to capture interdependencies. The contextualized categorical features are then concatenated with continuous features and processed through the MLP for final prediction. This design leverages the strengths of both contextual learning for categorical data and traditional MLP benefits for continuous data. Additionally, TabTransformer incorporates masked language modeling and replaced token detection, enabling it to pre-train on large amounts of unlabeled data, thus improving performance in low-labeled data scenarios and making it effective for real-world applications.

Recent advancements in TabTransformer models, such as the self-supervised TabTransformer introduced by Vyas \cite{vyas2024deep}, further refine this architecture by leveraging MLM in a pre-training phase to learn from unlabeled data. This self-supervised approach enhances the model’s ability to generalize by capturing intricate feature dependencies through self-attention mechanisms. By combining Transformer layers with MLP for final prediction, the model handles mixed data types and smaller dataset sizes effectively. However, trade-offs exist while the model demonstrates strong performance gains, particularly in semi-supervised settings, the reliance on masked language modeling pre-training increases computational overhead, potentially limiting scalability. Interpretability remains moderate, with attention scores providing insights into feature importance, though the model is less interpretable than traditional models like GBDT.

Another significant advancement is the GatedTabTransformer, introduced by Cholakov and Kolev \cite{cholakov2022gatedtabtransformer}, which enhances the original TabTransformer by incorporating a gated multi-layer perceptron. This modification improves the model’s ability to capture cross-token interactions using spatial gating units. The GatedTabTransformer boosts performance by approximately 1 percent in AUROC compared to the standard TabTransformer, especially in binary classification tasks. However, this comes at the cost of increased computational complexity due to the additional processing required for the spatial gating units. While the model shows improved performance, its scalability, and interpretability remain limited compared to simpler models like MLPs or GBDTs.

Therefore, while TabTransformer models offer notable improvements in handling tabular data through attention mechanisms and hybrid architectures, they present trade-offs in terms of performance, scalability, and interpretability. Recent variations like the self-supervised TabTransformer and GatedTabTransformer demonstrate the potential of these models to outperform traditional approaches, though at the cost of higher computational demands.

\subsection{FT-Transformer}
The FT-Transformer model, as presented by Gorishniy et al. \cite{gorishniy2021revisiting}, introduces a novel approach to addressing the challenges inherent in tabular data by leveraging attention mechanisms, hybrid architectures, and transformer-based methodologies. The model adapts the attention mechanism, originally designed for tasks like NLP, to process tabular data. In this context, the attention mechanism allows the model to capture complex relationships between heterogeneous features, including both numerical and categorical data as shown in Figure 11. By using attention to dynamically prioritize certain features, the model effectively models interactions that are often difficult to detect in traditional tabular data approaches.

\begin{figure}[htp]
    \centering
    \includegraphics[width=0.98\linewidth]{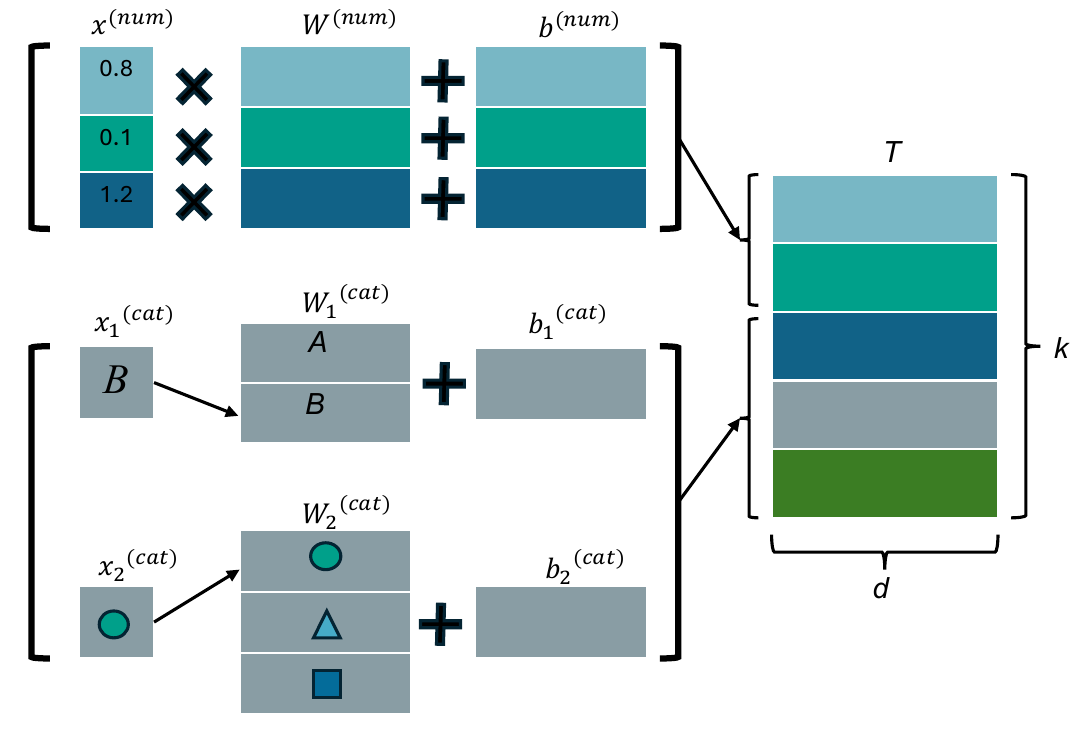}
    \caption{FT-Transformer Architecture \cite{gorishniy2021revisiting}}
    \label{fig:FT-Transformer}
\end{figure} 

In addition to attention, the FT-Transformer employs a hybrid architecture that integrates feature tokenization. This process transforms both numerical and categorical features into embeddings, which are then processed through layers of the Transformer architecture. The result is a model that is highly flexible and capable of handling diverse types of tabular data, a crucial advantage for tasks where tabular data can vary widely in feature types and distributions. This hybrid design bridges traditional feature encoding methods with the robust learning capabilities of Transformer-based approaches, enabling better generalization across different datasets.

Recent studies have demonstrated the effectiveness of the FT-Transformer across various applications. In the domain of heart failure prognosis, the FT-Transformer outperformed traditional models like Random Forest and Logistic Regression by capturing the non-linear interactions between medical features, such as demographic and clinical data \cite{kim2023predicting}. The use of attention mechanisms allowed the model to dynamically prioritize important health indicators, leading to more accurate predictions. Similarly, in intrusion detection systems, the FT-Transformer showed superior accuracy in identifying network anomalies by processing the highly structured nature of network traffic data \cite{saraniya2024securing}. The hybrid architecture seamlessly integrated categorical and numerical features, improving the model’s ability to detect both known and unknown threats. Additionally, advancements like stacking multiple transformer layers have been employed to further enhance the model’s capacity to capture long-range dependencies within the data, making it even more effective in complex tasks \cite{praveen2024enhanced}. While the FT-Transformer model demonstrates improved performance over other models, such as ResNet and MLP, particularly on various tabular tasks, it comes with certain trade-offs. In terms of interpretability, the model's complexity poses challenges. Traditional models like GBDT offer clearer interpretability, as their decision-making processes are more transparent. In contrast, the FT-Transformer’s reliance on attention mechanisms and deep layers makes it harder to explain, although the attention scores do provide some insight into feature importance. Furthermore, the model's scalability is another consideration; the computational demands of Transformer-based models, especially the quadratic scaling of the attention mechanism with the number of features, can become a limitation when applied to extensive datasets. Despite these limitations, the FT-Transformer's ability to generalize across diverse datasets makes it a promising model for tabular data analysis, offering significant advancements in predictive performance.

Building on these advancements, we present a performance and log-loss comparison between TabNet and FT-Transformer. As shown in Figure 12, the FT-Transformer consistently demonstrates superior performance as the number of random search iterations increases, while the log-loss for both models decreases at different rates. This comparison highlights FT-Transformer's enhanced generalization capabilities over TabNet, particularly in larger search spaces. While this figure provides an illustrative example of performance differences, unlike the previous survey on tabular deep learning \cite{borisov2022deep}, we have not offered a comparison of all tabular deep learning models, as a comprehensive evaluation across multiple models and diverse datasets is beyond the scope of this current survey. Future research should aim to conduct more extensive performance evaluations to thoroughly examine the strengths and limitations of these models.

\begin{figure}[htp]
    \centering
    \includegraphics[width=0.98\linewidth]{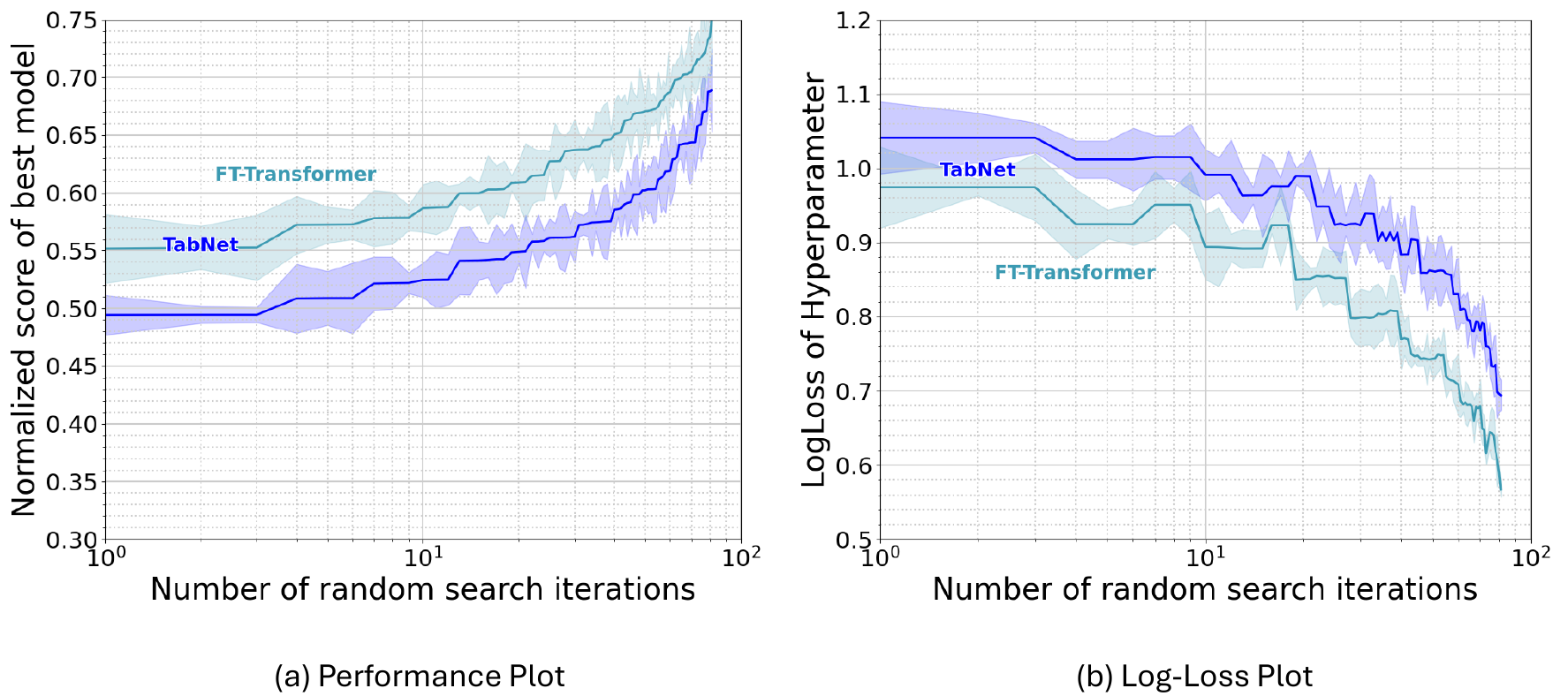}
    \caption{Performance and Log-Loss of TabNet and FT-Transformer Models}
    \label{fig: TabNet and FT-Transformer Comparison}
\end{figure} 

\subsection{DeepGBM}
The DeepGBM model represents an innovative approach to addressing the challenges of tabular data in deep learning, leveraging a combination of advanced techniques such as attention mechanisms, hybrid architectures, and knowledge distillation \cite{ke2019deepgbm}. While the model does not explicitly employ traditional attention mechanisms, it incorporates feature importance from GBDT, a method that allows the model to prioritize certain features over others. This process mimics attention by directing the model's focus to the most informative features rather than treating all inputs equally. By emphasizing the most relevant features, DeepGBM enhances its ability to handle both sparse categorical and dense numerical data, a crucial requirement in tabular data tasks.

Recent advancements in tabular deep learning further underscore DeepGBM’s role in combining neural networks with GBDT to achieve improved performance. In particular, the model’s hybrid architecture utilizes CatNN to handle sparse categorical features through embeddings and factorization machines, and GBDT2NN to convert the outputs of GBDT into a neural network format optimized for dense numerical features \cite{chen2019deep}. Figure 13 shows the structure of DeepGBM. This integration allows DeepGBM to leverage the strengths of both model types, overcoming limitations in traditional approaches that struggle to process mixed feature types in a unified framework.

\begin{figure}[htp]
    \centering
    \includegraphics[width=0.98\linewidth]{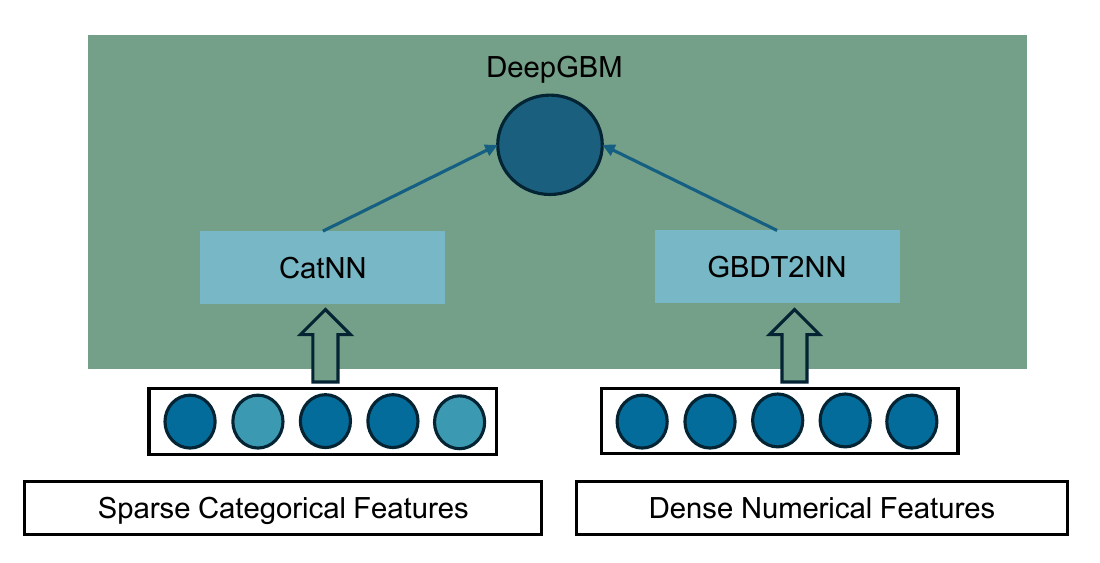}
    \caption{A DeepGBM Framework \cite{ke2019deepgbm}}
    \label{fig:DeepGBM}
\end{figure}

Though DeepGBM does not directly implement transformer models, it adopts ideas from transformer-based architectures, particularly in the form of knowledge distillation. By distilling the knowledge gained from GBDT trees into a neural network, including not just the predictions but also the tree structure and feature importance, DeepGBM retains the benefits of GBDT while enhancing its learning capacity \cite{yu2021windtunnel}. This mirrors how transformers use distillation to simplify complex models while preserving performance. 

The trade-offs in DeepGBM between performance, interpretability, and scalability reflect broader challenges in tabular deep learning. DeepGBM achieves higher accuracy by combining GBDT and neural networks but sacrifices some interpretability, as the added complexity of the neural network component reduces the transparency typically associated with tree-based models. Scalability is also a challenge, as the neural network elements require greater computational resources. However, models like WindTunnel have shown that this approach can boost accuracy while maintaining some of the structural benefits of the original GBDT \cite{yu2021windtunnel}. These trade-offs must be carefully balanced depending on the application, as DeepGBM excels in performance and efficiency, particularly for large-scale and real-time applications.

\subsection{Deep Attention Networks for Tabular Data (DANets)}
In recent advancements in tabular deep learning, the DANets model leverages attention mechanisms, hybrid architectures, and transformer-based approaches to tackle the challenges specific to tabular data processing. One of the key innovations in DANets is the use of a dynamic feature selection process, where relevant features are identified and emphasized through learnable sparse masks \cite{chen2022danets}. This approach, based on the Entmax sparsity mapping, allows the model to selectively focus on the most important features at each stage of the network, enhancing its ability to abstract meaningful representations from the data. This mechanism is akin to attention mechanisms used in transformer models, though specifically tailored for the irregular and heterogeneous nature of tabular data. 

\begin{figure}[htp]
    \centering
    \includegraphics[width=0.98\linewidth]{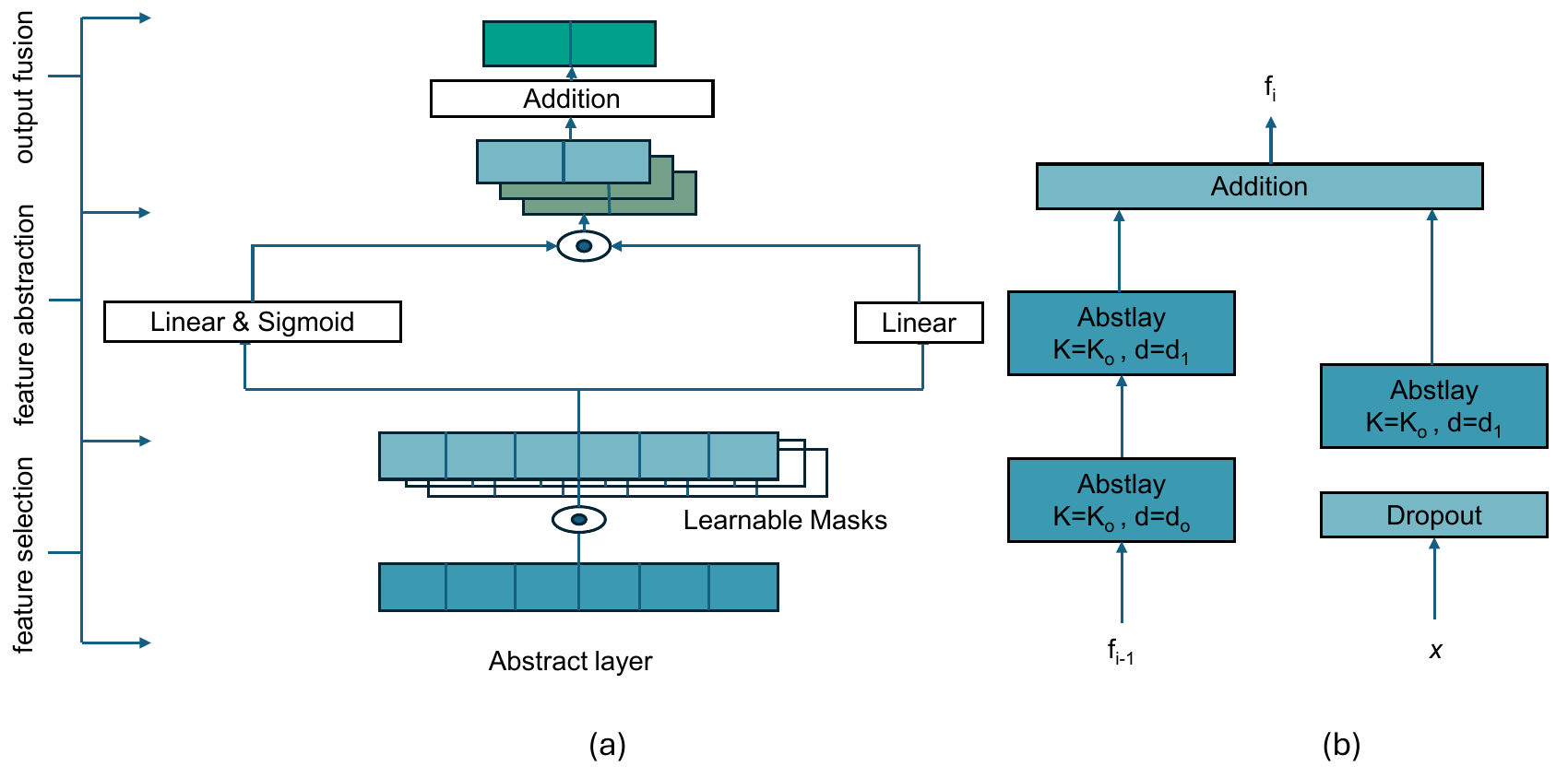}
    \caption{(a)DANets Abstract layer (b) An i'th Basic Block \cite{chen2022danets}}
    \label{fig:DANets}
\end{figure} 

DANets also incorporate hybrid architectures that blend feature grouping and hierarchical abstraction processes, similar to CNNs, but adapted for the unique structure of tabular data. The introduction of the Abstract Layer (ABSTLAY) as seen in Figure 14 enables the model to group correlated features and abstract higher-level representations through successive layers. Additionally, shortcut paths are employed, allowing raw features to be reintroduced at higher levels of the network, ensuring that critical information is retained and enhancing the model's robustness, particularly in deeper architectures. This design is similar to ResNet-style connections, where residual pathways prevent information loss and degradation in deeper networks, thus boosting performance.

DANets incorporate transformer-inspired ideas through the use of dynamic weighting and attention-like mechanisms, allowing the model to selectively focus on important features during the feature selection and abstraction processes. Although not a direct application of transformer models, these methods improve the handling of tabular data and boost performance, making DANets superior to traditional models like XGBoost and neural networks such as TabNet. However, this performance comes at the cost of reduced interpretability. While attention-based feature selection offers insights into the significance of specific features, the complexity of hierarchical abstraction obscures the decision-making process, making it less transparent than simpler models like decision trees. To address scalability, DANets utilize structure re-parameterization, which reduces computational complexity during inference, allowing deeper networks without overwhelming computational costs. Despite the performance boost from deeper architectures, the study notes that additional depth yields diminishing returns due to the limited feature space in tabular data.

\subsection{Self-Attention and Intersample Attention (SAINT)}
Recent advances in tabular deep learning have leveraged attention mechanisms and transformer-based approaches to address challenges in tabular data processing. The SAINT model leverages recent advances in tabular deep learning by integrating attention mechanisms, hybrid architectures, and transformer-based approaches to overcome the unique challenges of tabular data. SAINT uses two types of attention mechanisms: self-attention and intersample attention \cite{somepalli2021saint}. Self-attention allows the model to capture complex correlations between features within a single data sample, enabling it to model relationships that simpler models may miss. Intersample attention, a novel addition, enables the model to compare a row (data point) with other rows, allowing for a more dynamic learning process that adjusts based on patterns across the entire dataset. This mechanism proves useful in situations where some features may be noisy or incomplete, as the model can learn from other similar data points.

SAINT’s hybrid architecture combines both self-attention and intersample attention to create a comprehensive learning system. SAINT’s advanced architecture has also shown strong results in software defect prediction tasks \cite{mohapatra2024software}. By leveraging attention mechanisms and transformer-based approaches, SAINT effectively handles complex interactions between data points, improving defect prediction performance. It consistently outperforms traditional models like XGBoost and Random Forest, particularly when dealing with mixed data types. However, while SAINT offers improved accuracy, its complexity impacts interpretability due to the inclusion of intersample attention, making it less intuitive than simpler models. Additionally, the computational demands associated with SAINT’s attention mechanisms can pose scalability challenges, especially when working with larger datasets.

In addition to these innovations, SAINTENS, a modified version of SAINT, further enhances the model's ability to handle tabular data by addressing some of SAINT's limitations \cite{gutheil2022saintens}. SAINTENS employs the same attention mechanisms but includes a MLP ensemble to improve robustness when dealing with missing or noisy data. This approach, alongside contrastive pre-training and augmentation techniques such as Mixup and Cutmix, allows SAINTENS to generate stronger data representations, particularly in healthcare datasets where missing values are common. The trade-offs between these enhancements manifest in three key areas: performance, interpretability, and scalability. In terms of performance, SAINT and SAINTENS consistently outperform traditional machine learning models like GBDT and deep learning models like TabNet, especially when working with mixed feature types and datasets with limited labeled data. SAINT's attention mechanisms offer some degree of interpretability, allowing users to visualize important features and data points. However, the complexity introduced by intersample attention makes it less intuitive to interpret than simpler models. Lastly, while SAINT and SAINTENS scale well across large datasets, the computational demands of the attention mechanisms, especially intersample attention, can make these models more resource-intensive, particularly in larger datasets.

\subsection{Tabular BERT (TaBERT)}
The TaBERT model addresses the challenges of tabular data by incorporating attention mechanisms, hybrid architectures, and transformer-based approaches. A key innovation in TaBERT is its use of attention mechanisms, particularly the vertical self-attention mechanism, which operates over vertically aligned table cell representations across rows \cite{yin2020tabert}. This enables the model to capture dependencies between different rows and allows for a better representation of tabular data by focusing on relevant columns and rows in relation to a given natural language query. While this mechanism improves performance in handling tabular structures, it also introduces additional computational complexity, making it less scalable when dealing with very large datasets or tables containing numerous rows. Figure 15 illustrates the TaBERT architecture, which jointly processes natural language utterances and table schemas. It highlights how the model captures both text and tabular structures using multi-head attention and pooling mechanisms, enabling it to generate unified representations for downstream tasks like semantic parsing.

 \begin{figure}[htp]
    \centering
    \includegraphics[width=0.98\linewidth]{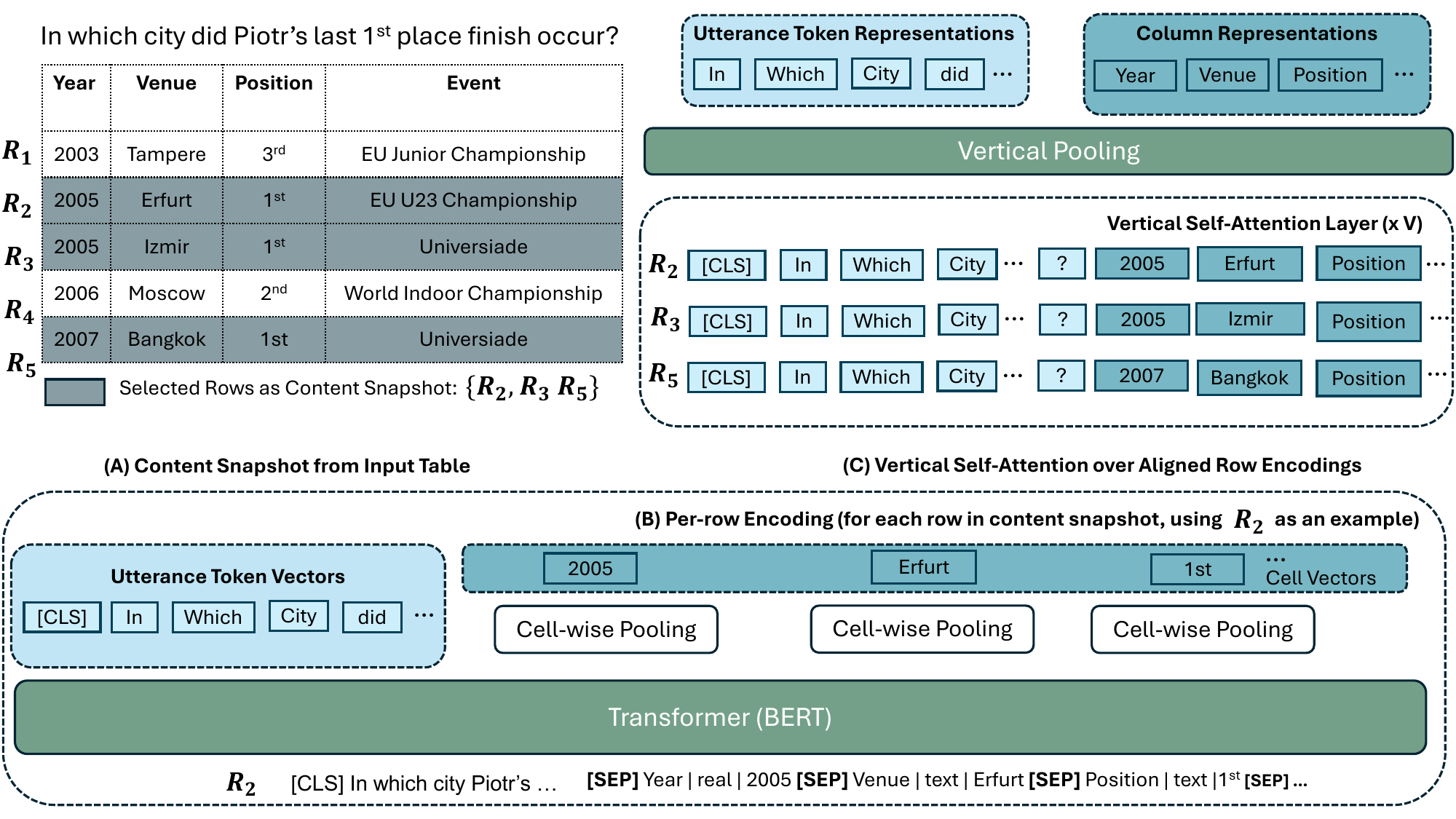}
    \caption{Overview of TaBERT's Method for Jointly Learning Representations of Natural Language Utterances and Table Schemas, Using an Example From WikiTableQuestions \cite{yin2020tabert}}
    \label{fig:TaBERT}
\end{figure}

In terms of architecture, TaBERT uses a hybrid approach known as content snapshots to reduce computational overhead. Instead of encoding all rows in a table, which would be costly, TaBERT selects a subset of rows that are most relevant to the natural language query. This allows the model to retain key information necessary for effective joint reasoning between text and tables while reducing the burden of processing unnecessary data. However, this comes with a trade-off: while content snapshots help scale the model to larger tables, there is a risk of losing critical information if the selected rows do not adequately represent the table’s full structure and content.

Built on a transformer-based pretraining framework, TaBERT benefits from learning representations of both natural language and structured data (tables). The model is pre-trained on a large corpus of 26 million tables and their corresponding text, using a BERT-like masked language modeling objective combined with table-specific objectives such as masked column prediction and cell value recovery. This pretraining improves the model’s ability to align textual and tabular data for downstream tasks like semantic parsing.

When evaluating performance versus interpretability, TaBERT excels in tasks like semantic parsing, where it outperforms models like BERT in benchmarks such as WikiTableQuestions as shown in Figure 15. However, the complexity introduced by the use of transformers and attention mechanisms makes TaBERT less interpretable than simpler machine learning models, such as decision trees, which offer more straightforward explanations for their decisions. In terms of scalability, the content snapshot mechanism helps the model handle larger tables more efficiently, but this comes with the risk of not fully capturing the table’s information.

\subsection{Tabular Transformer with Scaled Exponential Linear Units (TabTranSELU)}
The TabTranSELU model incorporates several recent advances in tabular deep learning, leveraging attention mechanisms, hybrid architectures, and transformer-based approaches to address the unique challenges of tabular data. One key innovation is the use of self-attention mechanisms, which allow the model to capture dependencies between different features in tabular datasets \cite{mao2024tabtranselu}. This self-attention approach is crucial for identifying relationships between input features, a task that can be particularly challenging with tabular data due to its lack of inherent structure, as seen in images or text. The attention mechanism computes scores by transforming the input into query, key, and value matrices, which enables the model to determine the weighted importance of different features. This helps the model learn inter-feature relationships more efficiently, ultimately improving its predictive performance. Figure 16 shows the input, transformer, and dense layer used in the TabTranSELU model. 

 \begin{figure}[htp]
    \centering
    \includegraphics[width=0.98\linewidth]{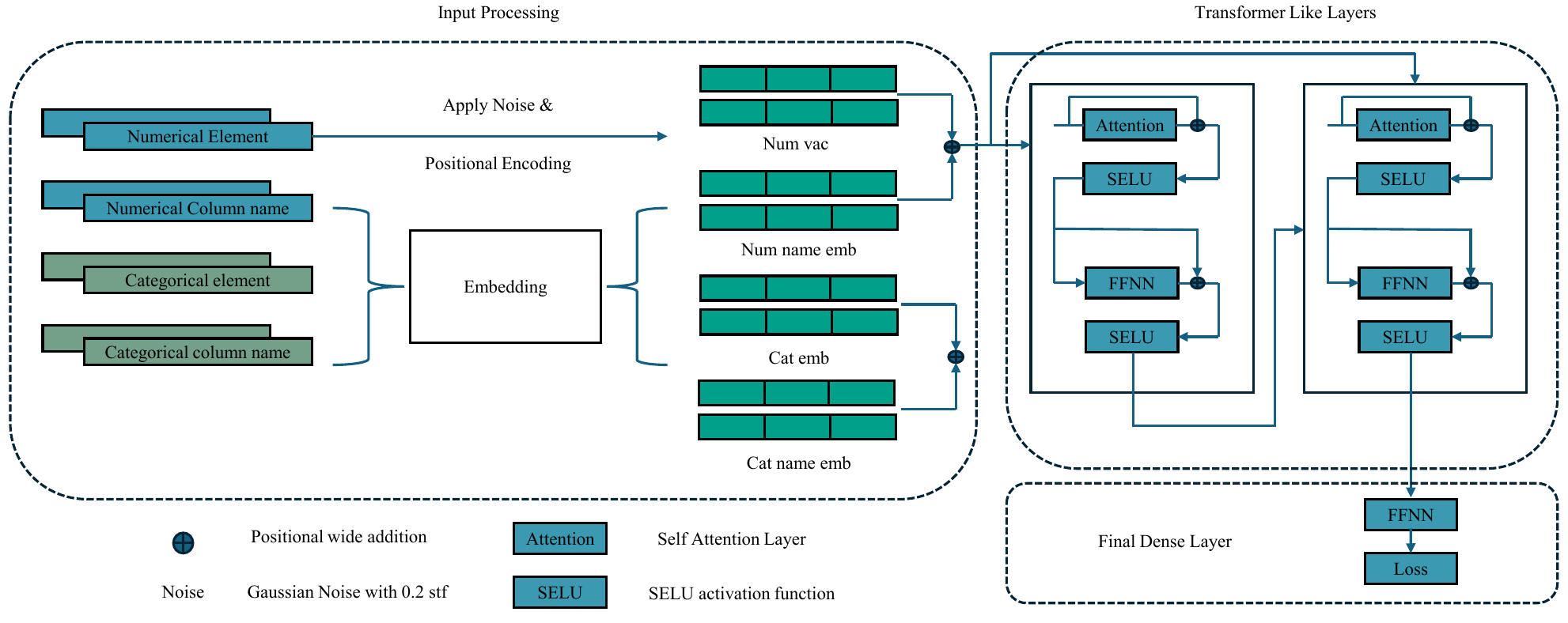}
    \caption{A TabTranSELU Framework \cite{mao2024tabtranselu}}
    \label{fig:TabTranSELU}
\end{figure}

The model also employs a hybrid architecture, adapting the traditional Transformer design for tabular data by simplifying its structure. Instead of utilizing the full stack of encoder and decoder layers, as seen in NLP tasks, TabTranSELU uses just a single encoder and decoder layer. This reduction in complexity helps tailor the architecture to the specific needs of tabular data without sacrificing performance. Moreover, the model integrates elements of both neural networks and transformers, allowing it to handle categorical and continuous features with equal efficiency. These features are processed separately through embedding layers, where categorical features are treated similarly to tokens in NLP, and numerical features undergo positional encoding to preserve their importance across different data instances.

One of the most significant adaptations of the TabTranSELU model is the replacement of rectified linear unit (ReLU) activations with scaled exponential linear units (SELU), addressing the "dying ReLU" problem, which is worsened by the presence of negative values in tabular data. SELU retains both positive and negative values, preventing the loss of latent information during training and making it more suitable for tabular datasets. Additionally, the use of positional encoding for numerical features preserves their order and significance, enhancing the model’s ability to handle continuous data. In terms of performance, TabTranSELU demonstrates competitive accuracy compared to traditional algorithms like gradient boosting decision trees (e.g., XGBoost), with only a slight 0.2 percent gap on larger datasets. It also performs well against similar transformer-based models, including TabTransformer, making it highly effective for predictive tasks, despite sacrificing a small amount of performance in exchange for broader functionality.

Interpretability is a key strength of the TabTranSELU model, with its embedding layers providing a clear understanding of relationships between features. Techniques like principal component analysis applied to the embeddings allow users to visualize how features and categories interact, offering valuable insights, especially when working with anonymized or unfamiliar datasets—insights that are often harder to achieve with traditional deep learning methods. In addition to interpretability, the model excels in scalability. By reducing the number of layers and incorporating SELU activations, it becomes more streamlined and less computationally intensive compared to traditional transformer models, making it well-suited for larger datasets and more efficient to train with minimal resource demand. Overall, TabTranSELU strikes an effective balance between performance, interpretability, and scalability, making it a strong choice for various tabular data applications. While we have already discussed several models from 2022 to 2024, it is important to note that a previous survey paper \cite{borisov2022deep} from 2022 did not include these more recent studies. The following section will explore the latest architectural innovations and models that push the boundaries even further, marking a new phase in the evolution of tabular deep learning.

\subsection{New Architectures and Innovations}
In recent years, the development of deep learning models for tabular data has accelerated, with new architectures emerging to tackle the growing complexity of this domain. Table 3 below highlights the key models introduced between 2023 and 2024, including innovative approaches such as LF-Transformer and ReConTab, which leverage advanced transformer-based and hybrid techniques to address challenges like feature interaction and noise. The table also outlines their architectures, training efficiencies, and notable features, offering a snapshot of the latest advancements in the field. LF-Transformer, for instance, employs both row-wise and column-wise attention mechanisms to capture complex feature interactions, using matrix factorization and latent factor embeddings to enhance prediction accuracy, particularly in noisy or incomplete datasets \cite{na2024lf}. This model excels in regression and classification tasks, though its complexity reduces interpretability and increases computational demands for larger datasets. Similarly, ReConTab utilizes a transformer-based asymmetric autoencoder to extract essential information from raw data, incorporating feature corruption techniques to enhance model robustness, though the added complexity results in higher computational costs and reduced transparency \cite{chen2023recontab}. GNN4TDL also builds on transformer-based autoencoder structures, leveraging feature corruption to improve robustness to noise and generalization, though it faces challenges in scalability and interpretability \cite{li2023graph}.

\thispagestyle{empty}
\begingroup
\fontsize{8pt}{8pt}\selectfont
\begin{longtable}{p{0.12\linewidth} p{0.2\linewidth} p{0.18\linewidth} p{0.3\linewidth}}
\caption{Timeline of DL Models for Tabular Data (2023-24)}
\label{tab:timeline_dl_models_2023_24} \\
\toprule
\textbf{\makecell{Model \\ (Year) \\ Source}} & 
\textbf{Architecture} & 
\textbf{\makecell{Training \\ Efficiency}} & 
\textbf{Main Features} \\
\midrule
\endfirsthead
\toprule
\textbf{\makecell{Model \\ (Year) \\ Source}} & 
\textbf{Architecture} & 
\textbf{\makecell{Training \\ Efficiency}} & 
\textbf{Main Features} \\
\midrule
\endhead
\midrule
\endfoot
\bottomrule
\endlastfoot
MambaTab (2024) \cite{ahamed2024mambatab} & Mamba block + final prediction layer & Supervised and self-supervised learning; High & Structured state-space models + feature incremental learning \\
\midrule
TP-BERTa (2024) \cite{yan2024making} & Transformer-based using relative magnitude tokenization + intra-feature attention & Supervised learning; Moderate & Transforms scalar numerical values into discrete tokens, integrates feature name-value pairs \\
\midrule
SwitchTab (2024) \cite{wu2024switchtab} & Asymmetric encoder-decoder & Self-supervised learning; Moderate to high & Employs asymmetric encoder-decoder structure to decouple mutual and salient features leveraging feature corruption \\
\midrule
CARTE (2024) \cite{kim2024carte} & Graph NN architecture using graph-attention layers & Self-supervised learning; Moderate & Transforms each row of tabular data into a graph representation \\
\midrule
BiSHop (2024) \cite{xu2024bishop} & Hopfield-based framework & Supervised learning; Moderate to high & Bi-directional sparse Hopfield modules to process tabular data column-wise and row-wise utilizing tabular embeddings for categorical and numerical features \\
\midrule
LF-Transformer (2024) \cite{na2024lf} & Column-wise transformer + row-wise transformer + latent factor embedding & Supervised learning; Moderate & Uses column-wise and row-wise attention, latent factor embeddings, matrix factorization to capture feature interactions \\
\midrule
TabTranSELU (2024) \cite{mao2024tabtranselu} & Self-attention + SELU activation + masked layer & Supervised learning; N/A & Applies positional encoding to numerical data and replaces ReLU with SELU activation \\
\midrule
TabR (2023) \cite{gorishniy2024tabr} & Feed-forward NN augmented with a retrieval-based mechanism & Supervised learning; Moderate & Integrates a retrieval-augmented mechanism using L2-based nearest neighbors with a feed-forward NN \\
\midrule
HYTREL (2023) \cite{chen2024hytrel} & Hypergraph structure-aware transformer (HyperTrans) & Self-supervised learning; High & Transforms tabular data into hypergraphs \\
\midrule
ReConTab (2023) \cite{chen2023recontab} & Asymmetric autoencoder & Self-supervised; semi-supervised; N/A & Uses transformer-based asymmetric autoencoder and feature corruption \\
\midrule
GNN4TDL (2023) \cite{li2023graph} & Graph neural network & Supervised learning; N/A & Transforms tabular data into graph structures using feature embeddings \\
\midrule
Trompt (2023) \cite{chen2023trompt} & Prompt learning & Supervised learning; Moderate & Uses prompt-inspired learning to derive sample-specific feature importances by combining column and prompt embeddings \\
\midrule
XTab (2023) \cite{zhu2023xtab} & Transformer based & Self-supervised learning; Moderate to high & Uses cross-table pretraining, data-specific featurizers, and embedding layers for categorical and numerical features \\
\end{longtable}
\endgroup

Expanding the scope of transformer models, MambaTab integrates structured state-space models with feature incremental learning, capturing long-range dependencies in tabular data more efficiently than standard self-attention mechanisms \cite{ahamed2024mambatab}. MambaTab’s ability to adapt to evolving feature sets enhances its scalability, but it sacrifices interpretability, lacking the attention mechanisms that explain feature importance in models like TabNet. SwitchTab employs an asymmetric encoder-decoder architecture that decouples mutual and salient features through separate projectors, improving feature representation in tabular data \cite{wu2024switchtab}. By using feature corruption-based methods, SwitchTab enhances performance and interpretability, but its complexity affects scalability, making it less efficient for very large datasets. Context Aware Representation
of Table Entries (CARTE) also utilizes advanced architectures, combining a Graph Neural Network (GNN) with graph-attention layers to represent each table row as a graphlet, enabling the model to capture complex contextual relationships across tables \cite{kim2024carte}. CARTE excels in transfer learning and performs well on heterogeneous datasets, although its graph-attention mechanisms reduce interpretability and scalability with large datasets.

In the realm of tokenization and prompt-based models, TP-BERTa stands out by applying Relative Magnitude Tokenization (RMT) to transform scalar numerical values into discrete tokens, effectively treating numerical data as words in a language model framework \cite{yan2024making}. Additionally, its Intra-Feature Attention (IFA) module unifies feature names and values into a coherent representation, reducing feature interference and enhancing prediction accuracy. However, this deep integration impacts interpretability compared to more straightforward models like gradient-boosted decision trees. Trompt employs prompt-inspired learning to derive sample-specific feature importance through the use of column and prompt embeddings, which tailor the relevance of features for each data instance \cite{chen2023trompt}. While Trompt boosts performance, especially for highly variable tabular datasets, the abstract nature of its embeddings compromises interpretability and adds complexity. 

Several other models combine innovative mechanisms with existing architectures to address tabular data challenges. TabR integrates a retrieval-augmented mechanism that utilizes L2-based nearest neighbors along with a feed-forward neural network, enhancing local learning by retrieving relevant context from the training data \cite{gorishniy2024tabr}. While this method significantly improves predictive accuracy, it introduces computational overhead during training, affecting scalability. BiSHop leverages Bi-directional Sparse Hopfield Modules to process tabular data both column-wise and row-wise, capturing intra-feature and inter-feature interactions \cite{xu2024bishop}. Its specialized tabular embeddings and learnable sparsity provide strong performance but at the cost of reduced interpretability and higher computational requirements, limiting its application to larger datasets.

Finally, Hypergraph-enhanced Tabular Data Representation Learning (HYTREL) addresses structural challenges in tabular data using a Hypergraph structure-aware transformer, representing tables as hypergraphs to capture complex cell, row, and column relationships \cite{chen2024hytrel}. This enables HYTREL to preserve critical structural properties and perform exceptionally well on tasks like column annotation and table similarity prediction, though the complexity of hypergraphs reduces interpretability. TabLLM, a novel approach, serializes tabular data into natural language strings to allow large language models (LLMs) to process it as they would with text \cite{hegselmann2023tabllm}. While effective in zero-shot and few-shot learning scenarios, TabLLM faces scalability issues and interpretability challenges due to the high computational demands of LLMs and their abstract representation of tabular data.

\section{Architectures and Techniques}
\subsection{Attention Mechanisms}

Attention mechanisms have become pivotal in enhancing feature selection, interpretability, and performance in various deep-learning models designed for tabular data. In models like TabNet, attention mechanisms focus on the most relevant features at each decision step, tailoring feature selection for each individual sample. This instance-wise feature selection improves efficiency and generalization, allowing the model to concentrate on the most critical features while minimizing distractions from less important ones \cite{arik2021tabnet}. Similarly, TabTransformer leverages self-attention layers to transform parametric embeddings into contextual embeddings, enabling the model to capture the dependencies between categorical features. This transformation allows for more refined feature selection, where the most relevant features are dynamically emphasized based on their interactions with others, resulting in improved performance across datasets \cite{huang2020tabtransformer}. Figure 17 further exemplifies this by demonstrating how Multi-Headed Self-Attention (MHSA) is applied between both features and samples in a tabular deep-learning model. By focusing attention first on the relationships between features and then between different samples, the model improves its ability to generalize and capture complex feature interactions, enhancing accuracy in tabular data processing.

 \begin{figure}[htp]
    \centering
    \includegraphics[width=0.98\linewidth]{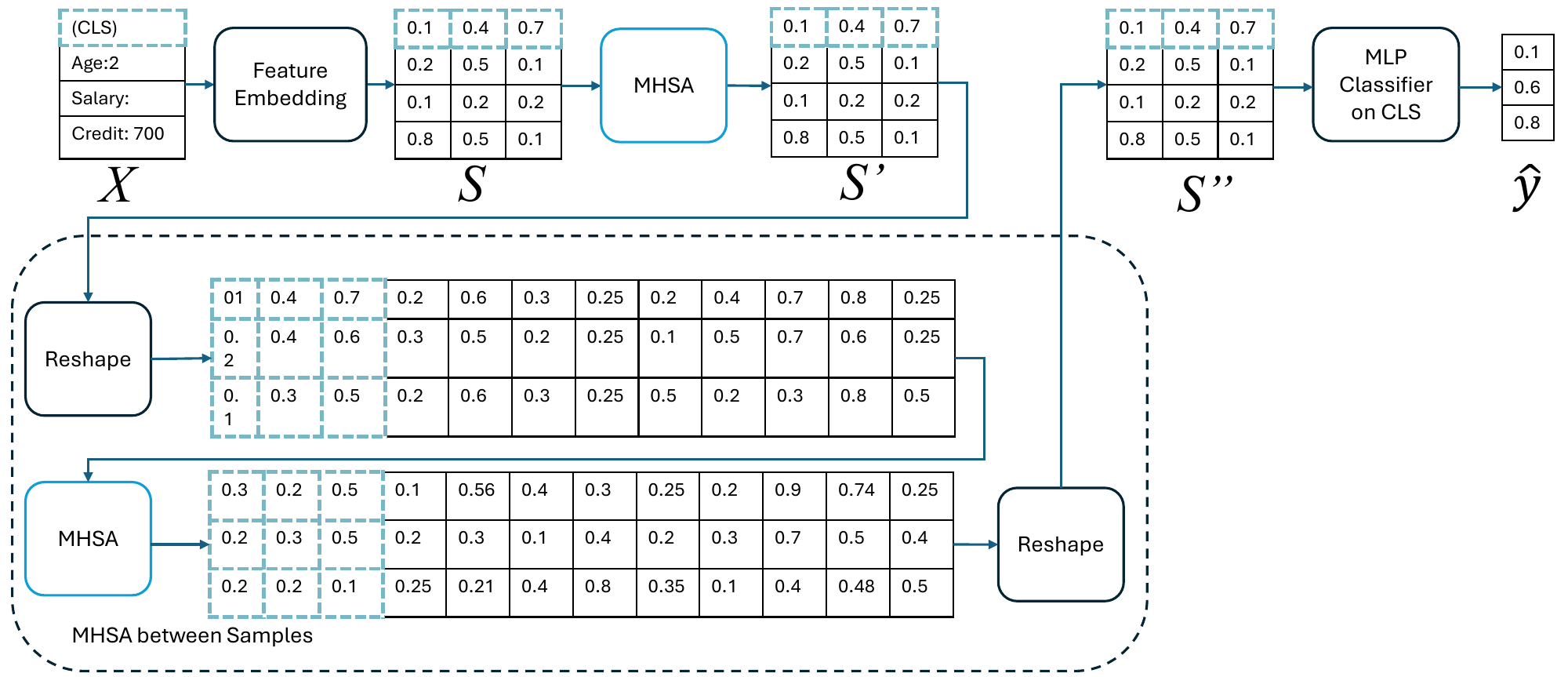}
    \caption{Feature and Sample Attention Using MHSA to Optimize Tabular Data Classification and Generalization \cite{rabbani2024attention}}
    \label{fig:Feature and Sample Attention}
\end{figure}

Building on this, SAINT introduces both self-attention and intersample attention mechanisms to further refine feature selection. The self-attention mechanism in SAINT focuses on interactions between features within a single data point, dynamically selecting important features based on their relationships \cite{somepalli2021saint}. This is similar to TabNet's emphasis on instance-specific feature selection but extends beyond by capturing deeper interdependencies between features, improving model adaptability and performance on heterogeneous datasets. SAINT's novel intersample attention adds another layer of sophistication by enabling data points to attend to other samples within the dataset. This allows SAINT to better handle noisy or missing features by borrowing relevant information from similar samples, a capability that is particularly useful in real-world datasets where data quality may vary. This cross-sample attention mechanism significantly enhances feature selection, making the model more robust to incomplete or corrupt data compared to traditional models like GBDTs and MLPs.

Both TabNet \cite{arik2021tabnet} and TabTransformer \cite{huang2020tabtransformertabulardatamodeling} offer significant advances in interpretability. TabNet operates at both local and global levels, enabling users to understand which features contribute to individual predictions while also providing a broader view of the overall model behavior. This transparency makes TabNet particularly useful in understanding model decisions for specific samples. Similarly, SAINT improves interpretability through its attention-based structure. In SAINT, attention maps highlight which features and samples are being prioritized during prediction, making it easier to trace the model's decision-making process and visualize feature importance. TabTransformer also enhances interpretability by generating contextual embeddings that cluster semantically similar features together in the embedding space. This clustering facilitates easier visualization and interpretation of feature relationships, making the model more transparent.

In terms of feature selection, TabNet integrates attention directly into the learning process, optimizing feature selection and model training simultaneously. Unlike traditional methods like forward selection or Lasso regularization, which apply uniform selection across the dataset, TabNet’s instance-wise selection adapts to the specific needs of each sample, resulting in more compact feature representations and a reduced risk of overfitting. InterpreTabNet, an improvement over TabNet, further boosts these capabilities with the MLP-Attentive Transformer and the Entmax activation function, leading to more precise feature selection \cite{wa2023stable}. Similarly, TabTransformer’s multi-head self-attention mechanism enables the model to dynamically capture feature interactions across the dataset. By attending to all other features, it efficiently selects the most critical ones while disregarding irrelevant data, which enhances the model's robustness against noisy or missing data. SAINT extends this concept by leveraging intersample attention, which allows features to interact across different samples. This mechanism not only improves feature selection but also provides a way for the model to learn from multiple data points simultaneously, enhancing its resilience to missing or noisy data. SAINT’s feature encoding method, which projects both categorical and continuous features into a shared embedding space, also outperforms traditional encoding techniques by allowing the model to learn from all feature types in a unified manner.

Both TabNet and TabTransformer, along with SAINT, showcase notable advancements in handling tabular data through their attention mechanisms, offering robustness, adaptability, and transparency. TabNet's attention-driven approach enhances gradient propagation and generalization, while TabTransformer excels in handling noisy and missing data, making both models suitable for real-world applications where data imperfections are common. SAINT builds on these strengths by introducing intersample attention, which allows the model to learn from relationships between samples, further enhancing its ability to handle complex data distributions. Additionally, TabTransformer and SAINT’s pre-training on unlabeled data in semi-supervised learning scenarios allows them to refine feature representations, contributing to improved performance when compared to models relying exclusively on labeled data.

\subsection{Hybrid Architectures}
Hybrid architectures, such as NODE and DeepGBM, leverage the strengths of both decision trees and neural networks to enhance generalization, capture complex feature interactions, and improve performance on tabular data. Both models capitalize on the interpretability and efficient feature splits provided by decision trees, while simultaneously benefiting from the gradient-based optimization and hierarchical representation learning typical of deep neural networks. This synergy between decision trees and neural networks allows these hybrid architectures to overcome the limitations faced by traditional models when dealing with tabular data, where deep learning models often underperform compared to shallower models like decision trees.

NODE employs differentiable Oblivious Decision Trees, a variant where all internal nodes of the same depth use the same splitting feature and threshold, which enables NODE to combine the inherent interpretability of decision trees with the backpropagation capabilities of neural networks \cite{popov2019neural}. This structure facilitates the learning of higher-order feature interactions by organizing decision trees into multi-layer architectures that mirror deep neural networks, thus enhancing generalization. Similarly, DeepGBM builds on this concept by incorporating two primary components: CatNN, which focuses on handling sparse categorical features, and GBDT2NN, which distills knowledge from GBDT into a neural network model to handle dense numerical features effectively \cite{ke2019deepgbm}. The GBDT2NN component takes advantage of GBDT’s ability to efficiently process numerical features while leveraging neural networks' flexibility in capturing complex feature interactions. In this way, both NODE and DeepGBM are able to represent intricate patterns in the data, improving performance on tasks traditionally dominated by simpler models like GBDT and boosting the effectiveness of tabular data predictions.

However, both architectures introduce challenges related to increased complexity. NODE's differentiable decision trees and multi-layer structures add computational overhead, making training more resource-intensive compared to simpler GBDT models. Similarly, DeepGBM’s distillation process, which involves learning leaf embeddings and managing multiple trees, introduces additional computational cost. Both models require careful hyperparameter tuning to optimize performance, which can make them more difficult to use in practice. Parameters like the number of layers, tree depth, tree groups, and output dimensions must be adjusted meticulously to avoid overfitting and ensure optimal learning. These complexities can increase the training time and resource demands for both NODE and DeepGBM, making them less efficient in terms of inference speed compared to their GBDT counterparts. Despite this, both models achieve comparable inference efficiency to GBDTs when implemented effectively, but the training process for NODE and DeepGBM tends to be longer due to the additional layers of differentiable optimization and knowledge distillation.

\subsection{Regularization and Optimization Techniques}
Kadra et al. \cite{kadra2021well} explored the effectiveness of regularization techniques such as Mixup, Dropout, and Weight Decay in mitigating overfitting in deep learning models for small tabular datasets. These methods enhance generalization by limiting weight magnitudes (Weight Decay), preventing neuron co-adaptation through random deactivation (Dropout), and generating synthetic data by interpolating between training samples (Mixup). While these techniques improve performance over traditional methods like GBDT, trade-offs include reduced interpretability and the need for careful hyperparameter tuning to maintain stability. The concept of a "regularization cocktail" combining multiple methods further demonstrates that well-regularized models can outperform both traditional and deep learning approaches on tabular data. Abrar and Samad \cite{abrar2022perturbation} also emphasize the role of Dropout and Weight Decay in combating overfitting in tabular datasets, particularly highlighting Dropout’s ability to improve generalization by forcing the network to learn diverse feature representations. They propose a novel periodic weight perturbation method, which prunes and regrows weights during training, achieving a balance between model compression and accuracy. This method outperforms traditional weight pruning by improving model generalization without the typical loss in accuracy, though it introduces challenges related to model interpretability due to the sparsity of the resulting models. Darabi et al. \cite{darabi2021contrastive} further demonstrate the effectiveness of Mixup in enhancing generalization, particularly through Contrastive Mixup, which interpolates samples in the latent space to avoid generating unrealistic data points. This improves the model’s stability and smoothes decision boundaries, but at the cost of reduced interpretability due to transformations in the latent space.

Shavitt and Segal \cite{shavitt2018regularization} take a different approach with Regularization Learning Networks, which assign different regularization coefficients to each weight based on feature importance. This allows for fine-tuned control over sparsity, reducing overfitting while maintaining interpretability. Regularization Learning Networks strike a balance between model complexity and stability, though they require sophisticated hyperparameter tuning. Similarly, Lounici et al. \cite{lounici2021muddling} introduce Muddling Labels for Regularization (MLR), a technique that uses label permutations and structured dithering to penalize memorization and improve generalization. MLR, tailored specifically for small tabular datasets, offers an effective alternative to Dropout and Weight Decay by maintaining model flexibility while reducing overfitting, though it may increase model complexity.

\section{Training Strategies}
\subsection{Data Augmentation}
Data augmentation techniques, such as SMOTE, GAN-based methods, and variational autoencoders (VAEs), have demonstrated varying degrees of effectiveness in improving the performance of deep learning models on tabular data, particularly in addressing class imbalance and small dataset issues. SMOTE, one of the classic techniques, has been widely used to oversample the minority class by generating synthetic samples \cite{machado2022benchmarking}. It does this by interpolating between existing data points in the feature space, which helps mitigate the class imbalance problem and can enhance model performance in imbalanced datasets. However, SMOTE performs well with categorical features, it struggles with continuous variables, as noted in experiments using datasets like Breast Cancer and Credit Card Fraud \cite{machado2022benchmarking}. The technique may struggle to maintain feature distributions when dealing with categorical data, leading to less realistic synthetic samples that may not fully capture the complexity of the original dataset. Wang and Pai \cite{wang2023enhancing} similarly note that SMOTE, although effective for initial data expansion, does not generate sufficiently diverse and realistic data, limiting its utility for more complex datasets.

GAN-based methods, particularly Conditional Tabular GAN (CTGAN) and Wasserstein GAN with Gradient Penalty (WCGAN-GP) have emerged as more advanced techniques for tabular data augmentation. These methods have demonstrated better performance than traditional techniques like SMOTE, especially when working with mixed-type tabular data containing both continuous and categorical features. Camino et al. \cite{camino2020minority} highlight the advantages of using GANs over SMOTE for minority class oversampling, emphasizing that GANs can generate more realistic and diverse samples. However, they also point out challenges specific to tabular data, such as difficulty in handling discrete outputs and mode collapse, where the GAN fails to generate a sufficiently varied dataset. Jeong et al. \cite{jeong2023bamtgan} introduce BAMTGAN, a variation of GANs, which incorporates a similarity loss to ensure the generated data maintains the original distribution and avoids mode collapse. Despite improvements, the challenge of balancing sample diversity and realism persists.

CTGAN addresses several challenges inherent to tabular data, such as handling non-Gaussian and multimodal distributions, by introducing mode-specific normalization and using a conditional generator to manage class imbalance Xu et al. \cite{xu2019modeling}. Sauber-Cole and Khoshgoftaar \cite{sauber2022use} offer a broad survey on the use of GANs to address class imbalance in tabular data. GANs are praised for generating realistic minority class samples and improving model performance on imbalanced datasets. However, challenges like mode collapse—where GANs fail to capture the diversity of the minority class and maintain realistic feature distributions, especially for categorical data, remain significant. Despite these issues, Sauber-Cole and Khoshgoftaar \cite{sauber2022use} highlight Wasserstein and Conditional GANs as promising solutions for overcoming these limitations. This allows CTGAN to generate more realistic and varied synthetic data while preserving the underlying data distributions. WCGAN-GP further improves the stability of GAN training by mitigating issues like vanishing gradients and mode collapse, which are common problems in standard GAN architectures. Compared to SMOTE, WCGAN-GP has been shown to produce synthetic data that better preserves data patterns and relationships, ultimately leading to better model performance and higher privacy protection \cite{mckeever2020synthesising}. Hybrid approaches that combine SMOTE with GAN-based methods address challenges faced by standalone models. Wang and Pai \cite{wang2023enhancing} introduced a hybrid model using SMOTE to augment small datasets, followed by WCGAN-GP to generate diverse and realistic synthetic data. This combination leverages SMOTE's statistical consistency and WCGAN-GP's ability to prevent overfitting, producing high-quality data while maintaining feature distribution, making it an effective solution for tabular data augmentation. 

VAEs are another promising approach for data augmentation, particularly for continuous data. VAEs regularize the latent space to generate smooth and realistic data distributions, and they have been effective in augmenting tabular datasets \cite{machado2022benchmarking}. However, they tend to struggle with mixed-type data and categorical features, where maintaining the original feature distribution becomes more challenging. Additionally, VAEs are prone to a phenomenon known as posterior collapse, where the latent space collapses into a narrow range, reducing the variability of the generated samples and leading to unrealistic outputs, particularly for minority classes.

One of the main challenges across these techniques is the difficulty in maintaining the original feature distributions, especially for continuous features and imbalanced categorical columns. While SMOTE works well for continuous data, it often falls short in handling categorical data. GAN-based approaches, such as CTGAN, use specific normalization techniques to address this issue, but even these advanced methods are not immune to challenges like mode collapse, where the model generates synthetic data lacking variability. GANs also require significant computational resources and careful tuning of hyperparameters to avoid these issues during training. Despite these challenges, GAN-based techniques, particularly WCGAN-GP, have demonstrated superior performance in generating high-quality, realistic synthetic data compared to traditional methods like SMOTE, making them a valuable tool for augmenting tabular datasets.

\subsection{Cross-validation}
Cross-validation is a crucial technique to ensure the generalization of deep learning models, especially for tabular data, where model overfitting and data imbalances can significantly affect performance. Richetti et al. \cite{richetti2023methods} emphasize the importance of cross-validation, particularly for smaller datasets where its role in preventing overfitting is more pronounced. In this context, k-fold cross-validation emerges as a popular method, with the authors applying an 8-fold cross-validation approach to achieve robust error measurements across different data partitions. Similarly, Zhu et al. \cite{zhu2021converting} applied tenfold cross-validation in their study on CNNs for tabular data transformed into image representations. The study highlights how k-fold cross-validation ensures generalization even after the transformation process, preventing overfitting, especially in limited or imbalanced datasets. Both studies underscore that while k-fold cross-validation offers robust performance evaluation, increasing the number of folds, such as from 5-fold to 10-fold, introduces higher computational costs without proportionally improving performance accuracy.

Wilimitis and Walsh \cite{wilimitis2023practical} provide a comparative analysis of cross-validation methods, focusing on the trade-offs between computational efficiency and model performance. They examine the commonly used 5-fold cross-validation alongside other variants like repeated k-fold cross-validation, finding that while more folds can slightly improve model evaluation, it also escalates computational demands. The study also explores nested cross-validation, a more unbiased method for performance estimation, particularly useful in healthcare models. However, the significant computational costs of nested cross-validation are highlighted due to its repeated training cycles during hyperparameter tuning. This mirrors the findings of Richetti et al. \cite{richetti2023methods}, who noted that methods like leave-one-out cross-validation (LOOCV) can be computationally impractical for larger datasets due to their repeated iterations.

Ullah et al. \cite{ullah2021explaining} expand on these ideas by discussing the use of stratified k-fold cross-validation, particularly for handling class imbalances in deep learning models for tabular data. By maintaining consistent class proportions in each fold, stratified cross-validation improves generalization, especially when working with imbalanced datasets, a key concern echoed in the previous two studies. Ullah et al. \cite{ullah2021explaining} also discussed the challenges of using LOOCV, which, despite providing unbiased performance estimates, comes with a high computational cost, especially for larger datasets. Nested cross-validation is similarly praised for its precision in mitigating data leakage during hyperparameter tuning but is noted for its quadratic time complexity, making it a computationally intensive choice.

These insights collectively suggest that while cross-validation techniques such as stratified k-fold and nested cross-validation are vital for improving the robustness of deep learning models on tabular data, they must be chosen carefully. The choice depends on balancing accuracy and computational efficiency, where simpler methods like k-fold cross-validation are more scalable, while more complex methods like nested cross-validation, although more precise, come with significant computational trade-offs.

\subsection{Transfer Learning for Tabular Data}
Transfer learning has demonstrated effectiveness in tabular data, particularly in addressing the limitations of small datasets. Pre-trained models, as emphasized by Levin et al. \cite{levin2022transfer}, enhance performance significantly when labeled data is limited, as these models transfer complex representations that outperform traditional models like GBDT. This effectiveness is most pronounced when there is alignment between the feature spaces of the upstream and downstream tasks, allowing models to generalize effectively across tasks. However, one of the key challenges, highlighted by Wang and Sun \cite{wang2022transtab}, is the inherent heterogeneity of feature spaces in tabular data. Tasks often involve different columns or feature types, making it difficult for pre-trained models to generalize without adaptation.

To address this challenge, several innovative methods have been developed. Levin et al. \cite{levin2022transfer} propose the use of pseudo-features, which allow models to manage mismatched or missing features during the transfer of knowledge across tasks with different feature sets. Similarly, Wang and Sun \cite{wang2022transtab} introduce the TransTab model, a transformer-based architecture that treats cells and columns as independent elements. This flexibility enables the model to handle tables with different formats, significantly improving generalization across tasks with varying feature types. Despite these advancements, issues such as catastrophic forgetting, where fine-tuning a pre-trained model on new data results in the loss of previously acquired knowledge, persist. Iman et al. \cite{iman2023review} address this by proposing progressive learning, a technique that adds new layers to pre-trained models during fine-tuning, preserving previously learned information while allowing the model to adapt to new tasks. Additionally, adversarial-based methods that use networks to extract transferable features across tasks have emerged as an effective strategy for improving model generalization in diverse tabular domains.

While these models represent significant progress, traditional methods like logistic regression and XGBoost still perform competitively with deep learning models in many tabular settings, as noted by Jin and Ucar \cite{jin2023benchmarking}. This is particularly true in cases where datasets differ significantly in feature types and distributions, further complicating the effectiveness of transfer learning. The challenge of imbalanced datasets, where models may skew performance toward majority classes, also remains a key issue.

Recent advancements focus on developing models tailored specifically for tabular data. Yan et al. \cite{yan2024making} introduced TP-BERTa, a pre-trained model that handles both categorical and numerical features using techniques such as relative magnitude tokenization and intra-feature attention. This model demonstrates improved performance over both deep learning and traditional methods by addressing the structural complexities of tabular data. Additionally, Jin and Ucar \cite{jin2023benchmarking} propose innovative architectures that leverage representation learning to facilitate knowledge transfer across tasks with similar feature types.

Pre-training and fine-tuning strategies, as noted by Levin et al. \cite{levin2022transfer}, remain central to improving performance in domain-specific applications. The use of self-supervised learning techniques, such as contrastive learning, has also proven effective, particularly in scenarios where labeled data is scarce. These methods enable models to learn useful features without extensive labeling, making them highly suitable for domains with limited labeled datasets. Moreover, El-Melegy et al. \cite{el2024prostate} introduced a novel approach that transforms tabular data into image-like formats, allowing the use of CNNs traditionally designed for image tasks. Coupled with GAN-based sampling, which generates synthetic data to balance datasets, this approach enables effective learning from small, sparse datasets. In summary, while transfer learning shows substantial promise in tabular data, its effectiveness is still hindered by challenges such as feature heterogeneity, catastrophic forgetting, and dataset imbalance. However, advancements like transformer-based architectures, progressive learning, and GAN-based data augmentation offer solutions to these challenges. As these methods continue to evolve, transfer learning will likely become a more robust and widely applicable tool for tasks involving tabular data.

\section{Future Directions}
As deep learning models continue to evolve for tabular data, two key areas stand out for future exploration: explainability and self-supervised learning. While current models offer impressive predictive capabilities, their lack of transparency remains a significant challenge in high-stakes fields like transportation engineering and healthcare. Enhancing model explainability and interpretability through advanced techniques like SHAP, LIME, and integrated gradients is essential for building trust and understanding in these models. Additionally, the growing field of self-supervised learning (SSL) offers significant potential to leverage vast amounts of unlabeled tabular data, improving model performance without relying on extensive labeled datasets. This section examines these promising directions and their potential impact on the future of tabular deep learning.

\subsection{Explainability and Interpretability}
Explainability techniques such as SHAP, LIME, and Integrated Gradients play a pivotal role in enhancing the interpretability of deep learning models, particularly in the domain of tabular data. However, their current implementations have limitations that necessitate further development, especially for real-world applications where trust and transparency are essential.

LIME has gained recognition for its ability to provide local explanations by creating simplified models around specific predictions. By perturbing the input data and observing the effects, LIME generates a local surrogate model that approximates the complex decision boundaries of the underlying deep learning model. Despite these strengths, LIME's reliance on kernel selection and its assumption of feature independence can lead to inconsistencies in high-dimensional datasets, as discussed by An et al. \cite{an2023specific}. 

SHAP, on the other hand, is grounded in game theory and provides a more globally consistent approach to feature contribution explanations. Unlike LIME, which focuses on local approximations, SHAP offers theoretically sound attributions of importance to each feature by calculating the marginal contribution of each feature to the prediction. Studies such as Ullah et al. \cite{ullah2021explaining} demonstrate that SHAP generally offers more accurate and consistent explanations than LIME. However, this comes at the cost of higher computational demands, which limits SHAP's practicality in real-time applications. In domains such as healthcare and finance, where regulatory compliance and trust are crucial, SHAP’s detailed and fair explanations have made it the preferred tool. Nevertheless, improving SHAP's computational efficiency without compromising its rigorous interpretability is crucial to making it viable for real-time and large-scale deployments.

Integrated Gradients provide a complementary approach, especially effective for models that involve multimodal data. Gao et al. \cite{gao2024improving} demonstrated the successful integration of SHAP and Integrated Gradients in a deep-learning model for hospital outcome prediction, using both clinical notes and tabular data. The combined use of these techniques allows for greater transparency by identifying the contributions of both text-based and structured data features. While this enhances trust among clinicians, the complexity of these explanations poses challenges for broader adoption. Simplifying these techniques to make them more accessible to non-technical users is necessary for their wider application in high-stakes environments such as transportation safety and healthcare.

Several studies emphasize the need for further refinement of these explainability techniques to improve their practical application. Dastile and Celik \cite{dastile2021making} applied SHAP in cancer prediction models and found that while SHAP enhanced the interpretability of the model, its computational demands made real-time application challenging. The authors suggest optimizing SHAP or developing more efficient explainability methods that retain interpretability while reducing resource consumption, especially in scenarios where real-time decision-making is critical. Similarly, Tran and Byeon \cite{tran2024predicting} used SHAP in a hybrid LightGBM–TabPFN model to predict dementia in Parkinson’s disease patients. SHAP provided valuable insights into the feature contributions, improving the model’s interpretability in a clinical setting. However, the study also highlights the need for further development of causality-driven explanations, integrating domain expertise to increase trust and applicability in medical environments. In summary, while SHAP, LIME, and Integrated Gradients have significantly improved the interpretability of deep learning models for tabular data, further development is needed to enhance their computational efficiency, stability, and accessibility for real-world applications where trust and transparency are crucial.

\subsection{Self-supervised learning}
SSL has been highly successful in domains like computer vision and NLP, where the inherent structures such as spatial relationships in images or semantic patterns in text make it easier to design effective pretext tasks. However, applying SSL to tabular data presents a unique set of challenges due to the lack of such explicit structures. Several studies have focused on adapting SSL techniques to tabular data to improve model performance and tackle the issues surrounding the design of meaningful pretext tasks and the effective use of unlabeled data.

One of the primary challenges in applying SSL to tabular data is the difficulty in designing effective augmentations and pretext tasks. Unlike structured data such as images, which exhibit spatial consistency, or text, which benefits from semantic coherence, tabular data lacks these natural structures. As a result, traditional augmentations like rotation in vision or token masking in NLP are not directly applicable. Wang et al. \cite{wang2024survey} discussed how SSL for tabular data needs to move beyond such tasks to design new methods that can capture the implicit relationships between features. Pretext tasks such as predicting missing values or reconstructing corrupted features have been proposed to overcome this limitation. Ucar et al. \cite{ucar2021subtab} introduce the SubTab framework as shown in Figure 18, which divides tabular features into multiple subsets and trains models to reconstruct features from these subsets, providing a novel multi-view approach for learning representations. This multi-view method implicitly acts as data augmentation, helping the model generalize better across different datasets. This is similar to Hajiramezanali et al. \cite{hajiramezanali2022stab}, who introduce the STab model, which avoids input-level augmentations by applying stochastic regularization techniques at various layers of the neural network, generating different views of the same data to improve the robustness of the learned representations.

\begin{figure}[htp]
    \centering
    \includegraphics[width=0.98\linewidth]{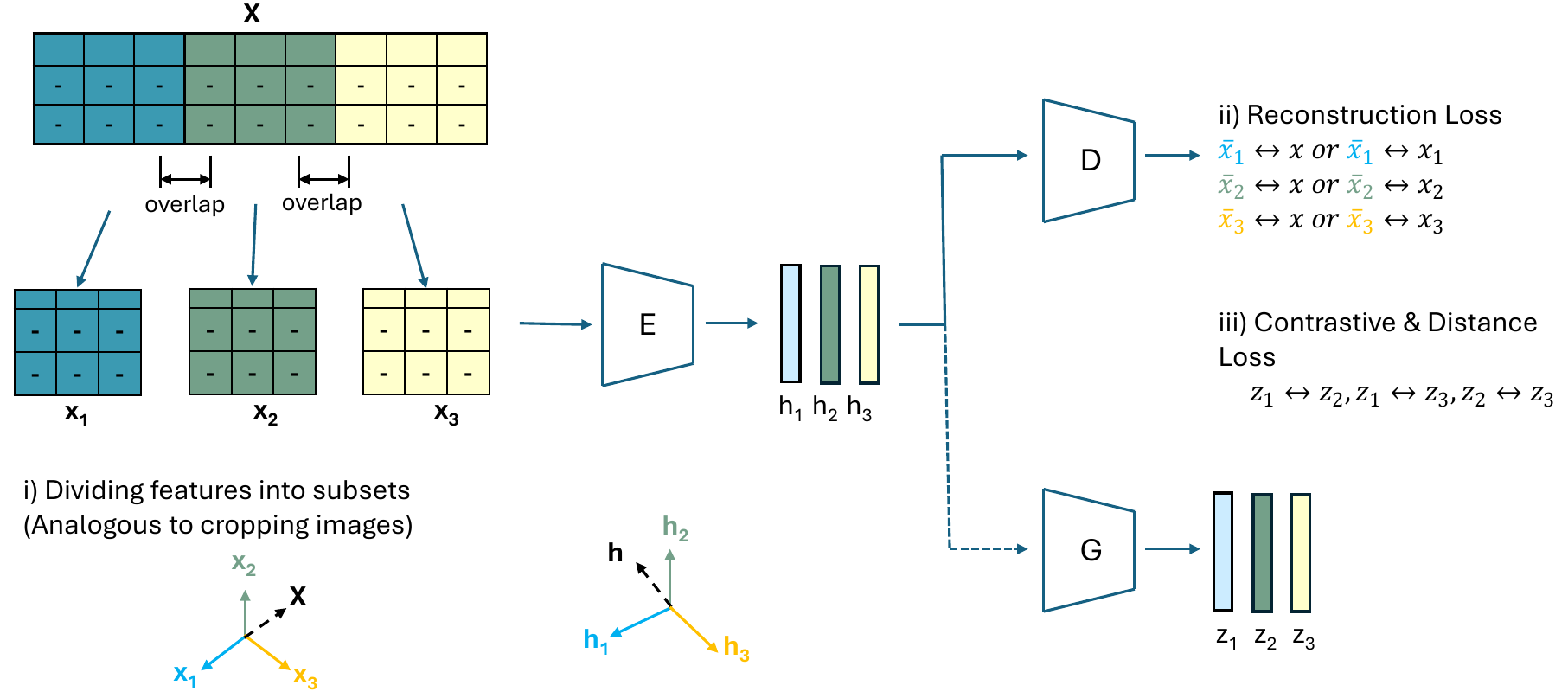}
    \caption{SubTab framework \cite{ucar2021subtab}}
    \label{fig:SubTab}
\end{figure} 

Chitlangia et al. \cite{chitlangia2022self} employed a different method, Manifold Mixup, which creates interpolations between hidden states rather than directly manipulating the input data. This method generates perturbed representations, allowing the model to recover original inputs and effectively handle high-cardinality features, thereby enhancing model performance without relying on manually labeled data. Similarly, Vyas \cite{vyas2024deep} applies a TabTransformer model, which uses self-attention mechanisms to capture dependencies between categorical and numerical features. By leveraging unlabeled data, this model learns effective representations without needing heavy reliance on labeled data, improving generalization across tasks.

Another approach to improving SSL’s efficacy on tabular data is through the careful design of reconstruction-based tasks that leverage feature subsets. Zheng et al. \cite{zheng2023tabular} apply the SubTab framework, where different subsets of features are used to reconstruct the full input, helping address the issue of heterogeneity in tabular data, where not all features contribute equally to the prediction task. This approach is echoed in VIME \cite{yoon2020vime}, which introduces two pretext tasks—feature vector estimation and mask vector estimation—that focus on reconstructing original data from masked and corrupted versions. These pretext tasks help the model learn robust representations by encouraging it to handle missing or noisy data effectively. Similarly, masked encoding for tabular data \cite{majmundar2022met} builds on this by incorporating masked encoding inspired by transformer models and using adversarial training during the reconstruction process. This adversarial component forces the model to recover features even in the presence of perturbations, making the learned representations more robust.

All these studies highlight the importance of leveraging unlabeled data in SSL, particularly in scenarios where labeled data is scarce. Leveraging large amounts of unlabeled data through techniques such as consistency regularization, as seen in VIME and MET, can significantly enhance model generalization even with limited labeled data. Wang et al. \cite{wang2024survey} emphasized that SSL techniques must make effective use of unlabeled data to ensure transferability across tasks, especially since tabular data often originates from diverse sources with varying feature distributions. This aligns with the findings of Chitlangia et al. \cite{chitlangia2022self}, where Manifold Mixup’s use of latent space perturbations helps generate meaningful augmentations without relying on input-level transformations. In SubTab, multi-view learning enables the model to capture different perspectives of the data, extracting more robust representations from unlabeled data. 

\section{Conclusion}
This survey reviewed the progress in deep learning models designed for tabular data, traditionally a challenging domain for deep learning. While classical models like GBDTs have long dominated tabular data tasks, new architectures such as TabNet, SAINT, and TabTransformer have introduced attention mechanisms and feature embeddings to better handle the complexities of heterogeneous features, high dimensionality, and non-local interactions. These models have made significant strides in enhancing interpretability and performance, with innovations like TabNet’s sequential attention and SAINT’s intersample attention, which dynamically capture relationships between features and rows of data.

However, challenges remain, particularly regarding computational efficiency and the risk of overfitting on smaller datasets. While models like TabTransformer and SAINT are computationally intensive, techniques like Mixup, CutMix, and regularization methods have been developed to address overfitting. Recent advancements, including hybrid models like TabTranSELU and GNN4TDL, have expanded the range of applications in many research domains. IGTD has further enhanced how deep learning models can transform tabular data into more structured formats for better performance.

One limitation of this survey is the lack of a detailed performance comparison across different models and datasets. Future research should focus on conducting more rigorous evaluations of tabular deep learning models on diverse datasets to gain a deeper understanding of their relative strengths and weaknesses. Alongside performance comparisons, further studies should aim to enhance the scalability and adaptability of these models, particularly in handling smaller or noisier datasets. Techniques such as transfer learning and self-supervised learning hold promise, as they allow models to benefit from large amounts of unlabeled data. Additionally, improving model interpretability and reducing computational costs will be crucial to broadening the applicability of deep tabular learning across industries like healthcare, finance, transportation, and infrastructure.

\begin{acks}
We extend our sincere gratitude to Khaled Aly Abousabaa for his assistance in preparing the images for this study. We also thank our colleagues at Texas State University for their valuable guidance throughout this work.
\end{acks}



\end{document}